\RequirePackage{fix-cm}
\documentclass[twocolumn]{svjour3}          
\smartqed  
\usepackage{graphicx}
\usepackage{float}
\usepackage{subfigure}
\usepackage{amsmath}
\usepackage{url}
\usepackage{multirow}
\usepackage{makecell, multirow, tabularx}
\usepackage{adjustbox}
\usepackage{tabularx}
\usepackage{xspace}
\usepackage{booktabs}
\usepackage{color}
\usepackage{amssymb,mathrsfs,amsmath}
\usepackage{arydshln}
\usepackage{amsfonts}
\usepackage{hyperref}
\newcommand{\eg}{\textit{e}.\textit{g}.}
\newcommand{\etal}{\textit{et al}.}
\newcommand{\ie}{\textit{i}.\textit{e}.}
\newcommand{\etc}{\textit{etc}}
\newcommand{\aka}{\textit{a.k.a.}}

\definecolor{hollywoodcerise}{rgb}{0.96, 0.0, 0.63}
\definecolor{lasallegreen}{rgb}{0.03, 0.47, 0.19}
\definecolor{hanpurple}{rgb}{0.32, 0.09, 0.98}
\definecolor{green(pigment)}{rgb}{0.0, 0.65, 0.31}

\usepackage{hyperref}
\hypersetup{colorlinks,linkcolor={red},citecolor={hollywoodcerise},urlcolor={red}}

\definecolor{codegreen}{rgb}{0,0.6,0}
\definecolor{codegray}{rgb}{0.5,0.5,0.5}
\definecolor{codepurple}{rgb}{0.58,0,0.82}
\definecolor{backcolour}{rgb}{0.95,0.95,0.92}

\begin{document}
\title{Priors in Deep Image Restoration and Enhancement: A Survey}

\author{Yunfan~Lu$^*$, Yiqi~Lin$^*$, Hao~Wu$^*$, Yunhao~Luo, Xu~Zheng, Hui Xiong, and~ Lin~Wang
\thanks{$^*$Equal  contribution, Corresponding author: Lin Wang}}

\institute{
Y. Lu and Y. Lin, H. Wu, X. Zheng are with the Artificial Intelligence Thrust, The Hong Kong University of Science and Technology (HKUST) Guangzhou, China. E-mail: ylu066@connect.hkust-gz.edu.cn,
linyq29@gmail.com,
hwubx@connect.ust.hk,
zhengxu128@gmail.com. Y. Luo is with Brown University. E-mail: devinluo27@gmail.com.
H. Xiong and L. Wang are with the Artificial Intelligence Thrust, HKUST Guangzhou, and Dept. of Computer Science and Engineering, HKUST, Hong Kong SAR, China. E-mail: \{xionghui, linwang\}@ust.hk
\\
\\
Preprint. Under review.
}

\date{Received: date / Accepted: date}

\maketitle

\vspace{-20pt}
\begin{abstract}
Image restoration and enhancement is a process of improving the image quality by removing degradations, such as noise, blur, and resolution degradation. Deep learning (DL) has recently been applied to image restoration and enhancement. Due to its ill-posed property, plenty of works have explored \textbf{priors} to facilitate training deep neural networks (DNNs).
However, the importance of priors has not been systematically studied and analyzed by far in the research community.
Therefore, this paper serves as the first study that provides a comprehensive overview of recent advancements of priors for deep image restoration and enhancement.
Our work covers five primary contents: (1) A theoretical analysis of priors for deep image restoration and enhancement; (2) A hierarchical and structural taxonomy of priors commonly used in the DL-based methods; (3) An insightful discussion on each prior regarding its principle, potential, and applications; (4) A summary of crucial problems by highlighting the potential future directions, especially adopting the large-scale foundation models as prior, to spark more research in the community; (5) An open-source repository that provides a taxonomy of all mentioned works and code links.
\end{abstract}

\section{Introduction}
\label{sec:introduction}

\begin{figure*}[t!]
\centering
\includegraphics[width=\textwidth]{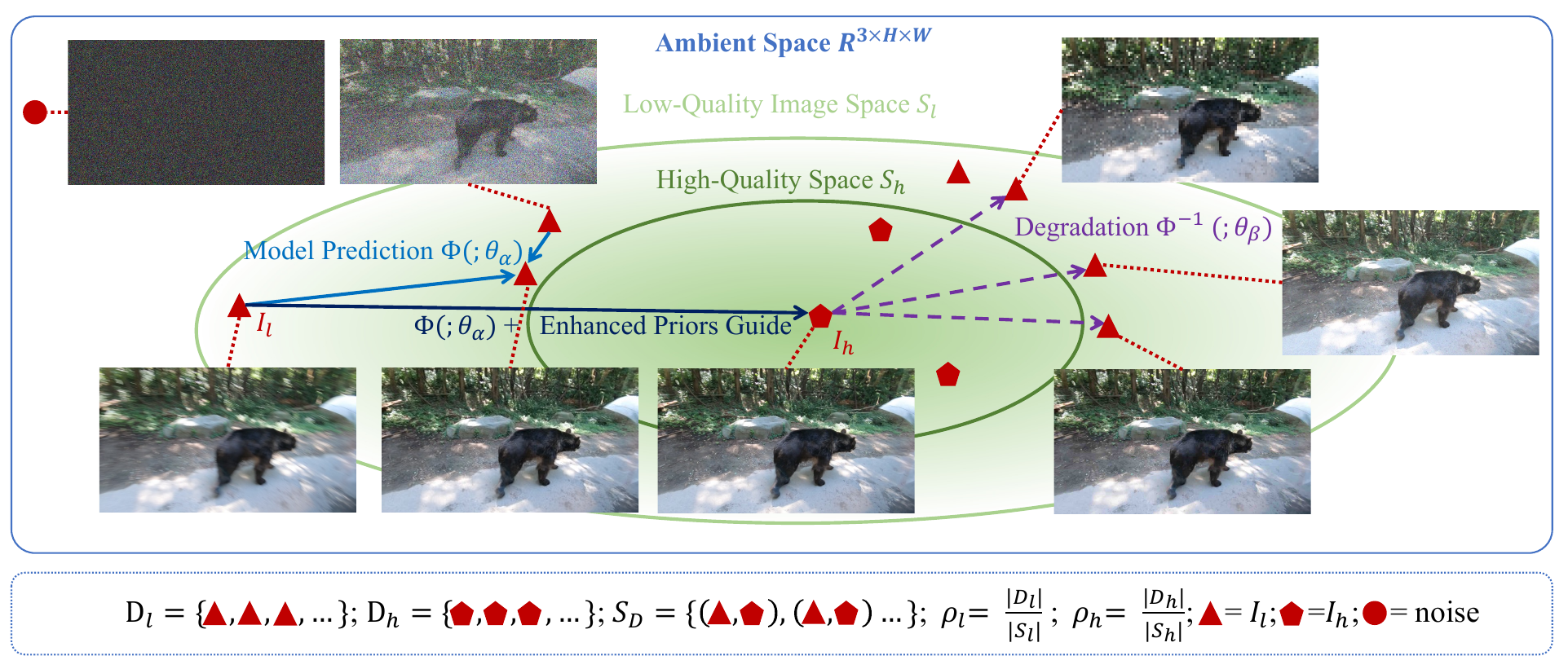}
\caption{The process of mapping the low-quality image set to the high-quality image set. The green oval represents the collection of low-quality images $S_l$. The dark green oval represents the collection of high-quality images $S_h$. Triangles $I_l$ represent sample points in the collection of low-quality images collection. Pentagons $I_h$ represent sample points in the collection of high-quality images collection. Circles represent meaningless points in the ambient space. $D_l$ is a low-quality dataset whose elements are samples of the low-quality image collection. $D_h$ is a high-quality dataset whose elements are samples of the high-quality image collection. $\rho_h$ is the sampling density of the high-quality images. $\rho_l$ is the sampling density of the low-quality images.}
\label{7-Backgorund}
\centering
\end{figure*}

Image quality often deteriorates during the capture, storage, transmission, and rendering process. Typical degradations include blur, resolution degradation, noise, and other artifacts.
Image restoration and enhancement is a process that attempts to improve the image quality by removing these degradations while preserving the indispensable image characteristics, as shown in Fig.~\ref{7-Backgorund}.

\begin{figure}[t!]
\includegraphics[width=0.5\textwidth]{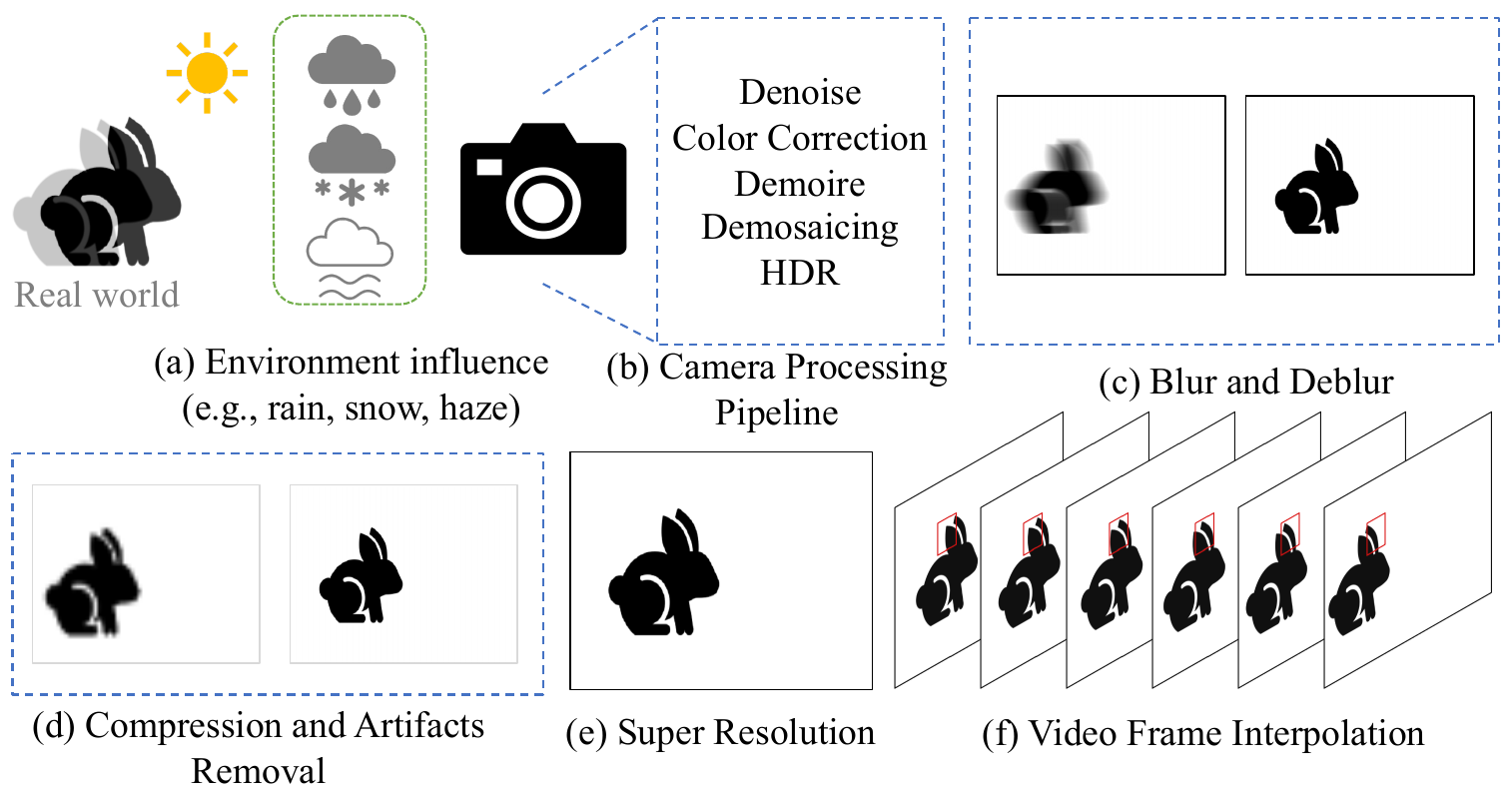}
\caption{Key steps in the photo imaging and transmission. (a) environment influences, which include rain, snow, and haze; (b) the camera processing pipeline; (c) blur and deblur; (d) compression artifacts removal; (e) super-resolution; (f) video frame interpolation.}
\label{0-OverallImageEnhancement}
\vspace{-10pt}
\end{figure}

Generally, image restoration and enhancement can be divided into six types, as shown in Fig.~\ref{0-OverallImageEnhancement}:
\textbf{(1) environmental influence removal}, which aims to remove the artifacts from images caused by environmental factors, including rain~\cite{jing2021hinet}, snow~\cite{liu2018desnownet}, haze~\cite{he2010single}, \etc.;
\textbf{(2) camera processing pipeline}, \aka, image signal processor (ISP), which converts the photoelectric signals into digital signals, including denoising~\cite{tian2020deep}, color correction~\cite{agarwal2006overview}, demoire~\cite{yang2020high}, demosaicing~\cite{kaur2015survey} and high dynamic range (HDR) imaging~\cite{wang2021deep}, \etc.;
\textbf{(3) deblurring}, which aims to restore sharp images by removing blur artifacts mainly caused by an object or camera motion~\cite{zhang2022deep};
\textbf{(4) compression artifacts removal}, which aims to reduce the low-quality problems, \eg, edge blur, blocking, and ringing artifacts, caused by lossy compression in the image storage and transmission process~\cite{svoboda2016compression};
\textbf{(5) super-resolution (SR)}, which aims to enhance the resolution of an image or video~\cite{wang2020deep};
\textbf{(6) video frame interpolation}, \aka, super slow motion, which aims to convert a low-frame-rate video into a high-frame-rate video~\cite{jiang2018super}.
Image restoration and enhancement enjoy many applications, \eg, human face recognition~\cite{oloyede2018improving}, autonomous driving~\cite{lee2021task}, and medical image analysis~\cite{islam2019image}.
Therefore, many early endeavors have actively explored the physical models to recover high-quality image content from its degraded version~\cite{he2010single}.

Recently, deep learning (DL) has brought new inspirations to image restoration and enhancement. DL-based methods usually achieve state-of-the-art (SOTA) performances on various tasks~\cite{zhong2020efficient,yang2020high,chan2021basicvsr++}.
Accordingly, diverse deep neural networks (DNNs) have been developed, such as convolutional neural networks (CNNs) ~\cite{zhang2017learning}, recurrent neural networks (RNNs)~\cite{zhong2020efficient}, generative adversarial networks (GANs) ~\cite{kupyn2018deblurgan} and vision transformers (ViTs)~\cite{liang2021swinir}, showing strong learning capability.
However, due to the highly ill-posed propriety of image restoration and enhancement, there still exist many challenges for the DL-based methods, \eg, heavily relying on large-scale data, hyper-parameter selection, and convergence instability.

\begin{figure*}[t!]
\centering
\includegraphics[width=\textwidth]{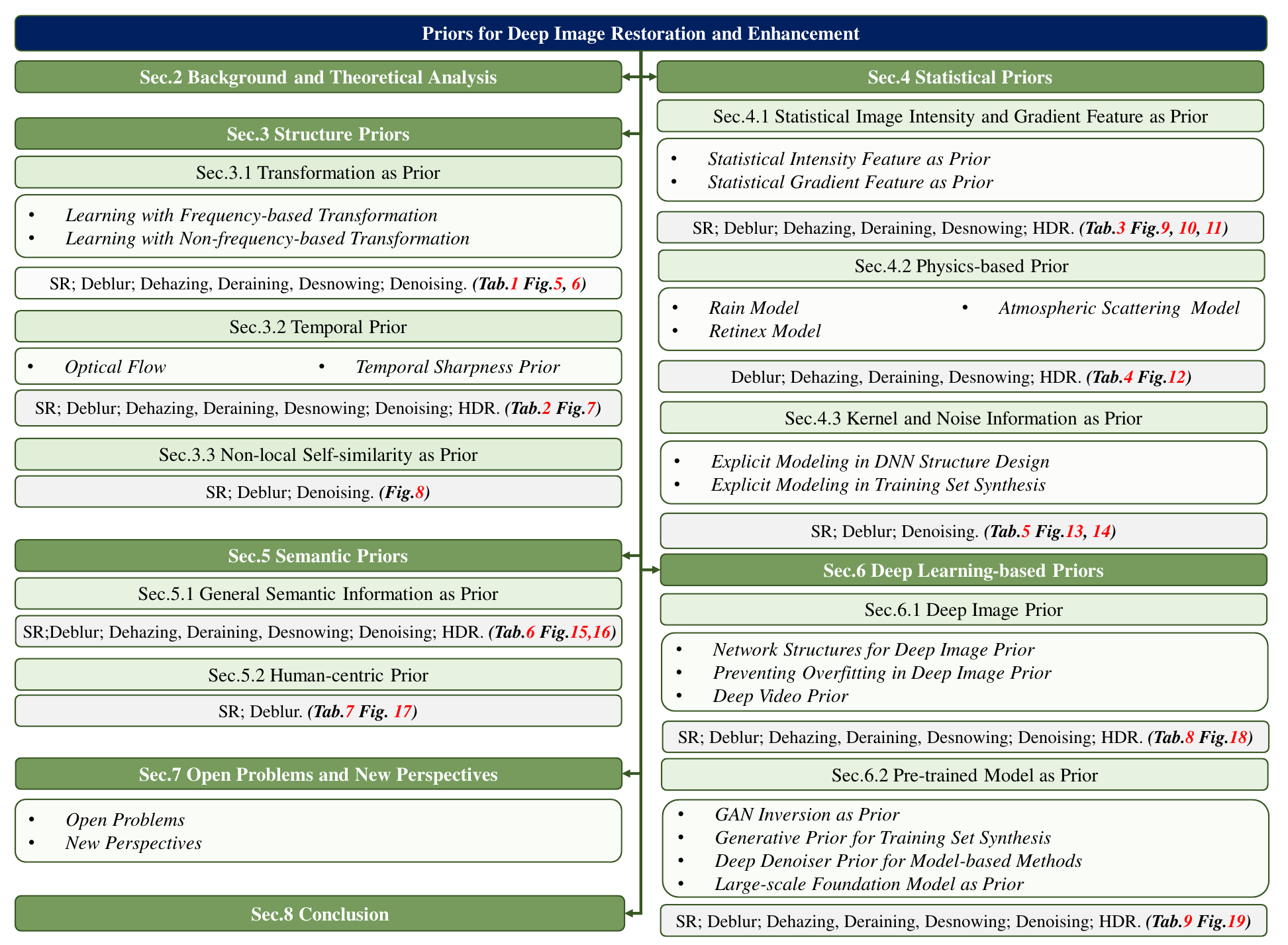}
\caption{Hierarchical and structural taxonomy of priors for deep image restoration and enhancement tasks. Applications of each prior, comparison tables of related papers, and viusal examples are shown in gray boxes.}
\label{2-ImagePrior}
\centering
\end{figure*}

For this reason, various priors have been explored for the acceleration of convergence and improvement of accuracy of DNNs in many image restoration and enhancement tasks~\cite{dhal2021histogram,kaur2020color}.
Specifically, priors refer to specific knowledge and patterns inherent to the image and video content.
Skillfully integrating these priors during network training or inference can accelerate convergence and significantly improve network performance~\cite{wang2021towards,he2010single}.
In this paper, we first present a systematic taxonomy of image enhancement and restoration priors into \textbf{four} primary categories: \textbf{(1) structure priors, (2) statistical priors, (3) semantic priors, and (4) deep learning-based priors.}
These four primary categories further encompass \textbf{ten} subcategories, representing a comprehensive collection of over \textbf{thirty} specific priors, as illustrated in Fig. \ref{2-ImagePrior}.
Note that most priors in the structure priors and statistical priors have been established in traditional image enhancement and restoration earlier to the emergence of DL-based techniques.
This paper emphasizes the application and potential of these priors within the context of the DL era.
In contrast, semantic priors and DL-based priors are inherently associated with learning approaches.

Although different specific priors are applied in different ways, we first attempt to use the manifold distribution~\cite{lei2020geometric} of data to explain the effectiveness of priors as a whole.
The distribution of the manifold reveals that natural images tend to cluster near a non-linear low-dimensional manifold within a high-dimensional image space $R^{H\times W\times 3}$, where $H$ and $W$ represent the image height and width, respectively, and $3$ corresponds to a color image, as shown in Fig.~\ref{7-Backgorund}.
The majority of images within the $R^{H\times W\times 3}$ space represent random noise rather than meaningful images.
The collection of meaningful images constitutes a subset of the $R^{H\times W\times 3}$ space.
The image priors serve as an effective means of describing specific knowledge within this subset of images.
For example, the dark channels prior~\cite{he2010single} describe a strong constraint for haze-free images.
Therefore, priors can effectively reduce the dimension of the prediction space (\ie, mapping from the degraded image space to the high-quality image space).
More mathematical principal analyses are provided in Sec.~\ref{sec:definition}.

This survey provides a comprehensive yet systematic overview of the recent developments of priors, \eg, large-scale foundation models as priors, in deep image restoration and enhancement.
Previously, some surveys, \eg,~ \cite{wang2021deep,zhang2022deep,wang2020deep,liu2021blind} only focused on a specific task in image restoration and enhancement.
For instance, Liu \etal~\cite{liu2021blind} proposed a taxonomy strategy to group the deep image super-resolution methods into different categories.
Wang \etal~\cite{wang2021deep} reviewed the crucial aspects of deep HDR imaging, \eg, domain input exposure, novel sensor data, and novel learning strategies.
Zhang \etal~\cite{zhang2022deep} discussed the common causes of image blurring and analyzed different CNN-based methods.
Differently, this survey considers a more general view by investigating different tasks to understand better what priors play in recent deep image restoration and enhancement methods.
Intuitively, we highlight the importance of priors that are advantageous in serving as guidance for diverse tasks in deep image restoration and enhancement. Accordingly, this survey serves as the \textbf{first} effort to offer an insightful and methodical analysis of the recent advances of priors, which may help spark new research endeavors in the community.

The main contributions of this study are five folds:
\textbf{(I)} We provide a comprehensive overview of priors in deep image restoration and enhancement, including four major parts:
structure priors (Sec. \ref{primary-prior:structure}), statistical priors (Sec. \ref{primary-prior:statistic}), semantic priors (Sec. \ref{primary-prior:semantic}), and deep learning-based priors (Sec. \ref{primary-prior:deep-learning}).
\textbf{(II)} We propose a general mathematical fundamental of the effectiveness of priors in image restoration and enhancement tasks based on the geometric theory (Sec.~\ref{sec:definition}).
\textbf{(III)} We analyze the connection among different priors and the application of each prior in different deep image restoration and enhancement tasks hierarchically and structurally (Sec.~\ref{sec:discussion}).
\textbf{(IV)} We discuss the open challenges of utilizing priors with deep image restoration and enhancement tasks and identify future directions to guide further research in this field, especially large-scale foundation models as priors (Sec.~\ref{sec:discussion}).
\textbf{(V)} We create an open-source repository with a taxonomy of all mentioned works and code links.
Our open-source repository will be updated regularly with new submissions in this research direction, and we hope it will shed light on future research.
The repository link is \href{https://github.com/VLIS2022/Awesome-Image-Prior}{GitHub Link}\footnote{\href{https://github.com/VLIS2022/Awesome-Image-Prior}{github.com/VLIS2022/Awesome-Image-Prior}}.
Meanwhile, we highlight and benchmark some representative priors on certain image restoration and enhancement tasks in Tab. {\color{hollywoodcerise}1}-{\color{hollywoodcerise}8}.
\textit{
Due to the lack of space, we show the experimental results of some representative priors for deep image restoration and enhancement tasks in the supplementary material.
}

In the following sections, we discuss and analyze various aspects of the priors in deep image restoration and enhancement tasks.
The remainder of this paper is organized as follows.
In Sec.~\ref{sec:definition}, we define the priors in deep image restoration and enhancement tasks and theoretically analyze the principles for different tasks.
In Sec.~\ref{primary-prior:structure} to Sec.~\ref{primary-prior:deep-learning}, we provide a comprehensive discussion of each primary prior and their corresponding subcategory priors.
In Sec.~\ref{sec:discussion}, we discuss the open problems and future directions.
In Sec.~\ref{sec:Conclusion}, we conclude this article.
\section{Background and Theoretical Analysis}
\label{sec:definition}
In this section, we provide the general mathematical fundamentals of the effectiveness of priors in deep image restoration and enhancement based on the geometric theory~\cite{lei2020geometric}.
We denote the relationship between a high-quality image $I_h$ and a low-quality image $I_l$ in Eq.~\ref{eq:low_quality_to_high_quality}, where $\Phi$ is a mapping function, and $\theta_{\alpha}$ is the parameter.
For deep image restoration and enhancement networks, $\Phi$ can be flexibly replaced by a DNN~\cite{zhong2020efficient}.
\begin{equation}
    I_h = \Phi(I_l; \theta_{\alpha})
    \label{eq:low_quality_to_high_quality}
\end{equation}

The degeneration function is the inverse of the restoration and enhancement process, as formulated in  Eq. \ref{eq:high_quality_to_low_quality}, where $\Phi^{-1}$ is the degeneration function, such as the blurring model, noising model, down-sampling model, and $\theta_{\beta}$ are its parameters.
\begin{equation}
\small
    I_l = \Phi^{-1}(I_h; \theta_{\beta})
    \label{eq:high_quality_to_low_quality}
\end{equation}

Based on the manifold distribution law~\cite{lei2020geometric}, we denote the ambient space as $\mathbb{R}^{C \times H \times W}$, where $C$ is the number of RGB channels, $H$ and $W$ are the height and width of an image, respectively.
Most images in the ambient space have no physical meaning, just like noise, as shown in Fig. \ref{7-Backgorund}, the red circle represents a meaningless image.
We are only interested in images with meaningful information, such as natural images, face images, and text images.
In the image restoration and enhancement task, we focus on the high-quality image set $S_h$ and the low-quality image set $S_l$.
They are all distributed in low-dimensional manifold spaces of the ambient space, as shown in Fig. \ref{7-Backgorund}.
The relationship between $\mathbb{R}^{C\times H \times W}$, $S_h$, and $S_l$ is shown in Eq. \ref{eq:set_relationship}, where $|\cdot|$ represents the cardinal number, \ie, the size of a set.
\begin{equation}
\small
    \left\{
         \begin{aligned}
            & \{S_h \cup S_l\} \subset \mathbb{R}^{ C\times H \times W} \\
            & |S_h| \ll |\mathbb{R}^{ C\times H \times W}| \\
            & |S_l| \ll |\mathbb{R}^{ C\times H \times W}|
         \end{aligned}
    \right.
    \label{eq:set_relationship}
\end{equation}
\vspace{-10pt}
\begin{equation}
\small
    D(x) = \int_{\Omega} c(x, \Phi(x;\theta_{\alpha})) dx
    \label{eq:transform_cost}
\end{equation}

A deep image restoration and enhancement model $\Phi(\cdot;\theta_{\alpha})$ describes a mapping from the low-quality image set $S_l$ to the high-quality image set $S_h$.
Training the mapping model $\Phi(\cdot ; \theta_{\alpha})$ is to find the minimum value of Eq. \ref{eq:transform_cost}, where $D(x)$ is the total cost of two distributions mapping, $c(\cdot,\cdot)$ is the cost function, such as $L_1, MSE$, \cite{lei2020geometric}, and $\Omega$ is a convex region to represent a manifold considering model $\Phi(\cdot;\theta_{\alpha})$ is a homeomorphic transformation.
Different cost functions lead to different mapping models.
Correspondingly, the degradation model $\Phi^{-1}(\cdot;\theta_{\beta})$ describes the opposite mapping.

Therefore, we argue that $p_\Phi$, the performance of the image restoration and enhancement model $\Phi(\cdot;\theta_{\alpha})$, depends on three aspects:
(1) $p_l$, the accuracy of the estimation of the input distribution according to $S_l$, \aka, the low-quality image distribution;
(2) $p_h$, the accuracy of the estimation of output distribution according to $S_h$, \aka, the high-quality image distribution;
(3) $f_\Phi$, the ability to map the input to the output.
$p_l$ and the $p_h$ are proportional to the sampling density $\rho$ when the sampling is uniform.
$f_\Phi$ can be measured by the mapping ability of the network.
The relationship between $p_{\Phi}$ and $p_l$, $p_h$, $f_{\Phi}$ can be simplified to Eq. \ref{eq:performance_relationship}.
\begin{equation}
    \small
    p_{\Phi} \propto p_l \times p_h \times f_\Phi
    \label{eq:performance_relationship}
\end{equation}

Assume that $S_D$ is an image restoration and enhancement dataset, which can be formulated by Eq. \ref{eq:dataset_set}. Here,
$D_l$ is the low-quality image dataset.
$D_h$ is the high-quality image dataset.
$N_l$ and $N_h$ are the counts of the datasets.
$I_{(l,i)}$ and $I_{(h,j)}$ are sampled points from the set $S_l$ and set $S_h$.
In general, a dataset is a collection of sampled points on the manifold space, as shown in Fig. \ref{7-Backgorund}.
The pentagons are points from a dark green ellipse, and the collection of pentagons is a high-quality image dataset.
The sampling density of low-quality and high-quality are shown in Eq. \ref{eq:density}, where $C$ is the size of the dataset, and $|S|$ is the cardinal number of the set.
Therefore, the sampling density of low-quality and high-quality images is shown in Eq. \ref{eq:sampling_density_low_high}.
In addition, the $S_l$ and $S_h$ are constant,  Eq. \ref{eq:performance_relationship} is equal to Eq. \ref{eq:performance_relationship_v2}.
\begin{equation}
\small
    \left\{
         \begin{aligned}
            & D_l = \{I_{(l,i)} | i \in [1, N_l]\} \\
            & D_h = \{I_{(h,j)} | j \in [1, N_h]\} \\
            & S_D = \{(I_l, I_h) | I_l \in D_l, I_h \in D_h\}
        \end{aligned}
    \right.
    \label{eq:dataset_set}
\end{equation}
\vspace{-12pt}
\begin{equation}
\small
    \rho = \frac{C}{|S|}
    \label{eq:density}
\end{equation}
\vspace{-12pt}
\begin{equation}
\small
    \left\{
         \begin{aligned}
            & \rho_l=\frac{N_l}{|S_l|} \\
            & \rho_h=\frac{N_h}{|S_h|}
         \end{aligned}
    \right.
    \label{eq:sampling_density_low_high}
\end{equation}
\vspace{-12pt}
\begin{equation}
\small
    p_{\Phi} \propto N_l \times |S_l|^{-1} \times N_h \times |S_h|^{-1} \times f_\Phi
    \label{eq:performance_relationship_v2}
\end{equation}

The effectiveness of priors is reflected in the impact on $p_l$, $p_h$, and $f_\Phi$.
Some priors can constrain high-quality image sets,
such as dark channel prior~\cite{he2010single}, gradient distribution~\cite{zhang2020deep}, $L_0$ regularity~\cite{zhang2021blind}.
These priors describe some properties of the high-quality image set $S_h$ and reduce the search space for the network to estimate the distribution of $S_h$, equivalent to increasing $\rho_h$.
Generally, priors can improve network performance by increasing one or more values of $p_l$, $p_h$, and $f_\Phi$.

\begin{figure}[t!]
\centering
\includegraphics[width=0.5\textwidth]{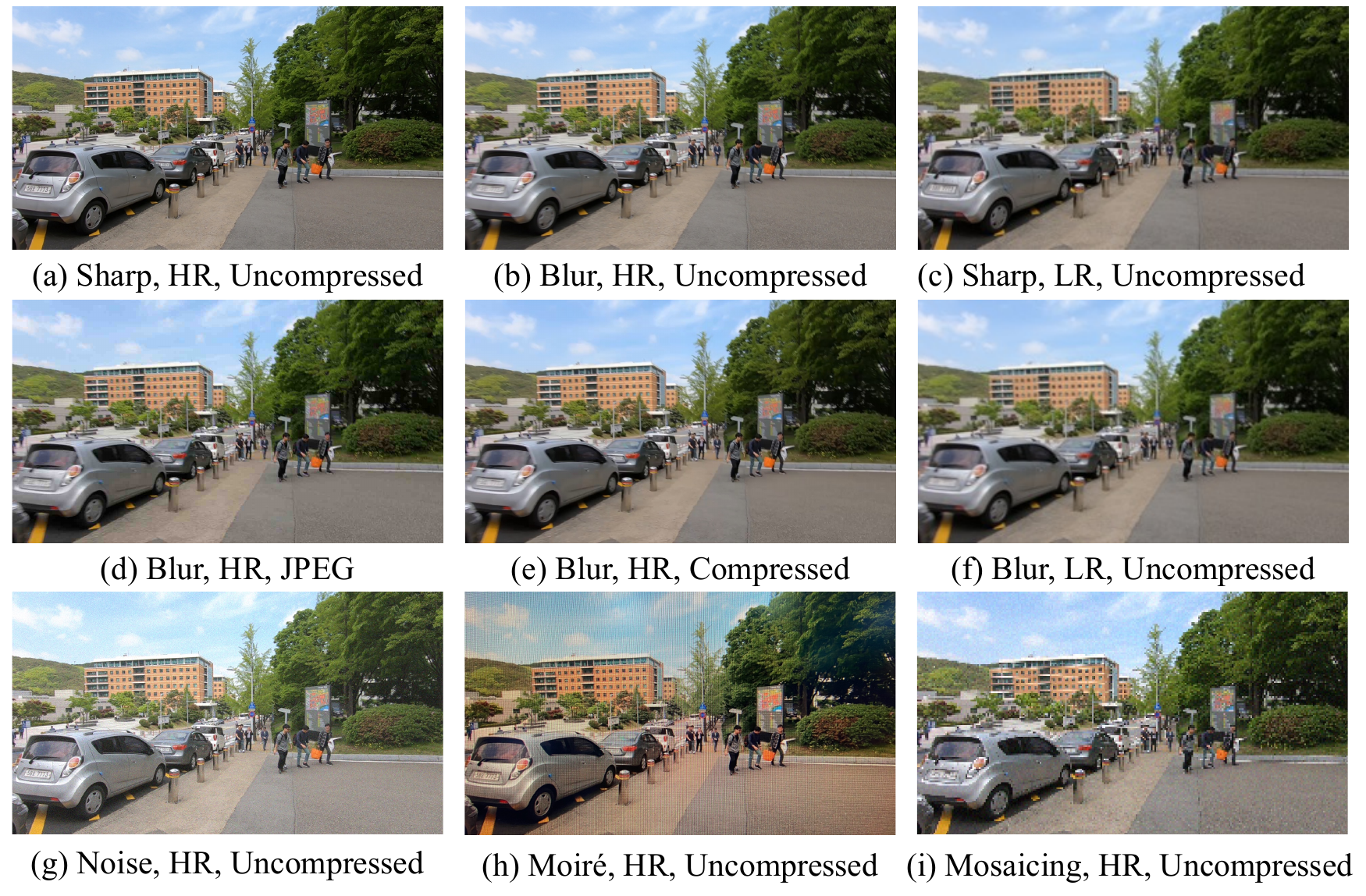}
\caption{Different kinds of degeneration artifacts. }
\label{fig:6-vis-all}
\centering
\end{figure}

In this survey, all the investigated priors can be analyzed by these principles.
Specifically, some priors focus on increasing $p_l$ by directly generating more low-quality images, \eg, modeling kernel and noise information (Sec. \ref{subsubsec:kernel_prior_2}) or using generative  models (Sec. \ref{subsubsec:pretrain2}) for the training set synthesis.
In addition, some priors focus on increasing $p_h$ by embedding the knowledge of high-quality images into DL frameworks, \eg, using statistical information (Sec. \ref{sec:statistical}) or deep denoiser (Sec. \ref{subsubsec:pretrain3}) as a regulation term, using GAN inversion (Sec. \ref{subsubsec:pretrain1}) to leverage the knowledge from latent space.
Moreover, some other priors focus on improving $f_\Phi$ by guiding the design of the network's structure, \eg, physical models (Sec. \ref{sec:physical}), temporal prior (Sec. \ref{subsec:temporal_prior}), modeling kernel and noise (Sec. \ref{subsubsec:kernel_prior_1}), non-local similarity (Sec. \ref{sec:non}) or introducing extra information to improve of DL models \eg, transformation prior (Sec. \ref{sec:transformation_as_prior}), semantic prior (Sec. \ref{sec:high_level_information_prior}), human-centric prior (Sec. \ref{sec:facial_prior}).
Moreover, deep image prior (Sec. \ref{sec:deep_image_prior})
takes advantage of DNNs structure $f_\Phi$ to increase $\rho_h$, so as to perform image restoration and enhancement from the single low-quality image.

\section{Structure Prior}
\label{primary-prior:structure}

Structure priors are based on the underlying structural \cite{shi1994good} and geometric characteristics~\cite{hoiem2005geometric} of the image or video, providing rules and constraints about the image and video reconstruction.
We consider transformation-based priors, temporal priors, and non-local self-similarity priors as part of the structure priors due to their ability to provide valuable insights into the image structure.

Transformation-based priors (Sec. \ref{sec:transformation_as_prior}) include frequ-ency-based and non-frequency-based transformations.
Frequency-based transformations focus on the frequency domain.
By analyzing and manipulating the frequency components of an image, these transformations can reveal important structural information and guide the restoration and enhancement process \cite{chen2021all,fuoli2021fourier,maggioni2021efficient,liu2021invertible}.
Non-frequency-based transformations, on the other hand, operate in the spatial domain and can include geometric transformations or other spatial manipulations.
These transformations can help preserve structural details and improve the overall image quality by leveraging the spatial relationships between pixels or regions \cite{yang2017deep,ren2020single,fang2020soft}.

Temporal prior (Sec. \ref{subsec:temporal_prior}), especially in video, leverages temporal coherence and consistency in frame sequences \cite{pan2020cascaded,liang2022vrt}.
By considering the dynamic structure changes between consecutive frames, temporal prior aids in preserving the temporal sharpness and reducing artifacts in video restoration and enhancement tasks.

Non-local self-similarity-based priors (Sec. \ref{sec:non}), another component of structure priors, exploit the inherent self-similarities present in images and frames.
By identifying and utilizing similar patches or regions within the image or frames, these priors help preserve the structural details and textures, contributing to the overall image restoration and enhancement \cite{wang2018non,zhang2019residual}.
Utilizing non-local self-similarity-based priors ensures that the restored image maintains structural coherence and consistency, resulting in enhanced visual quality.

\vspace{-10pt}
\subsection{Transformation-based Prior}
\label{sec:transformation_as_prior}

\noindent \textbf{Insight:} \textit{Transforming images to different domains can bring favorable properties for network training, e.g., some noise patterns are more apparent in specific frequency sub-bands.
}

\begin{table}[t!]

\centering
\caption{Application of transformation as prior in deep image/video restoration and enhancement tasks. $\star$ denotes `first work'. $I/O$ denotes priors applied to inputs and outputs. $F$ denotes priors applied to the feature map.}
\label{tab:freq}
\setlength{\tabcolsep}{2mm}{
\resizebox{\linewidth}{!}{
\begin{tabular}{ccc}
\hline
Method & Task & Transformation Types \\
\hline
\hline
CVPRW 2017~\cite{bae2017beyond} & SR;Denoising & \makecell[c]{Wavelet, $I/O$}\\
CVPRW 2017~\cite{guo2017deep} & SR & \makecell[c]{Wavelet, $I/O$} \\
ICCV 2017~\cite{huang2017wavelet} & SR & \makecell[c]{Wavelet, $I/O$} \\
CVPRW 2018~\cite{liu2018multi} & SR \& Denoising & \makecell[c]{Wavelet, $F$ $\star$} \\
TIP 2019~\cite{yang2019scale} & Deraning & \makecell[c]{Wavelet, $I/O$} \\
ICCV 2019~\cite{deng2019wavelet} & SR & \makecell[c]{Wavelet, $I/O$} \\
ECCV 2020~\cite{liu2020wavelet} & \makecell[c]{Demoireing \&\\Deraining} & \makecell[c]{Wavelet, $I/O$} \\
ECCV 2020~\cite{rong2020burst} & Denoising & \makecell[c]{Wavelet, $F$} \\
CVPR 2021~\cite{liu2021invertible} & Denoising & \makecell[c]{Wavelet, $I/O$} \\
CVPR 2021~\cite{maggioni2021efficient} & Denoising & \makecell[c]{Customized, $I/O$} \\
ICCV 2021~\cite{fuoli2021fourier} & SR & \makecell[c]{Fourier, $I/O$} \\
ICCV 2021~\cite{chen2021all} & Desnowing & \makecell[c]{Wavelet, $I/O$}\\
ICIP 2022 \cite{fan2022half} & \makecell[c]{Low-light} & \makecell[c]{Wavelet, $F$} \\
ICASSP 2022 \cite{ma2022wavelet} & \makecell[c]{Underwater \\image\\ enhancement} & \makecell[c]{Wavelet, $I/O$}\\
PR 2023 \cite{tian2023multi} & Denoising & \makecell[c]{ Wavelet, $F$} \\
\hline
\end{tabular}}}
\vspace{-5pt}
\end{table}

\begin{figure}[t!]
    \centering
    \includegraphics[width=0.47\textwidth]{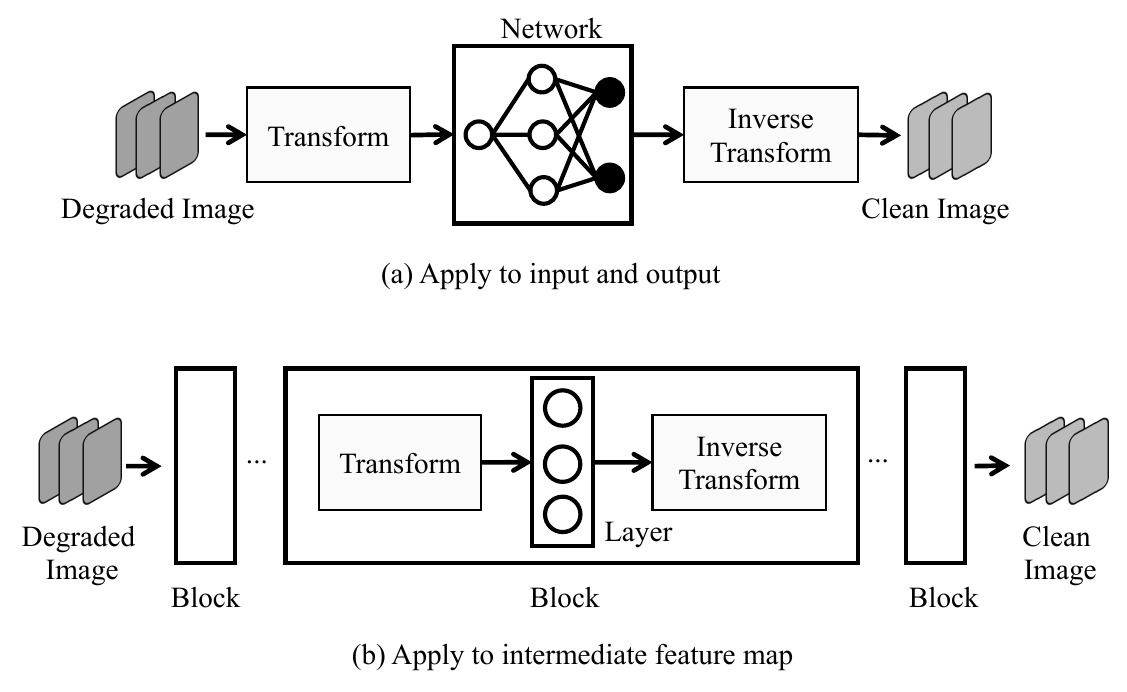}
    \caption{Illustration of using transformation as prior.}
    \label{fig:18-FreqPrior}
\end{figure}

In this section, we review the representative works that use transformation as prior to reduce the solution space for deep image restoration and enhancement.
Different from the data augmentation applying the transformation to improve the data diversity, using transformation as prior aims at leveraging extra information from the transformed domain to improve network model performance and training efficiency.
The transformations used as prior can be divided into two categories: frequency-based transformation~\cite{liu2018multi} and non-frequency-based transformation~\cite{yang2017deep2}.
The representative works of using the frequency-based transformation as prior for deep image restoration and enhancement are summarized in Tab.~\ref{tab:freq} and more details can be found in the supplementary material.

\begin{figure}[t!]
    \centering
    \includegraphics[width=0.47\textwidth]{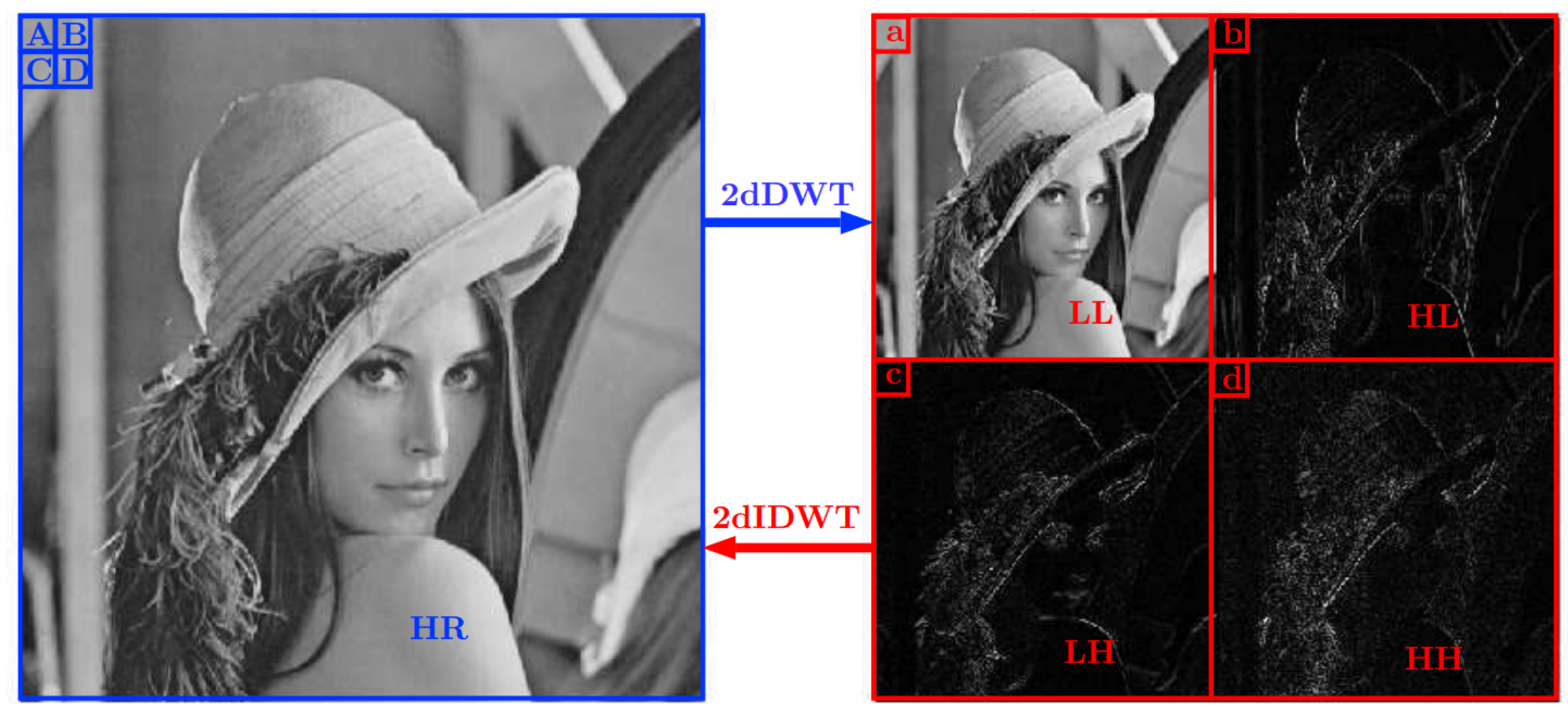}
    \caption{The 2D discrete
wavelet transform and its inverse transformation used in~\cite{guo2017deep}, which can be applied to the features or input for image enhancement and restoration tasks.}
    \label{fig:dwt}
\end{figure}

\subsubsection{Learning with Frequency-based Transformation}
Transforming data into the frequency domain, such as using Discrete Fourier Transform (DFT) or Discrete Wavelet Transform (DWT) as shown in Fig~\ref{fig:dwt}, allows the decomposition of data into different frequency sub-bans for component-wise analysis.
These approaches have been widely studied for image restoration~\cite{mallat1989theory,figueiredo2003algorithm} before the DL-based methods were developed.
In the context of deep image restoration, some works have empirically or experimentally shown that certain patterns, \eg, noise~\cite{bae2017beyond}, moire pattern~\cite{liu2020wavelet}, rain~\cite{yang2019scale}, haze~\cite{Fu_2021_CVPR} or snow~\cite{chen2021all} are more apparent in certain frequency sub-bands.
They can be useful priors to improve the network training efficiency by integrating the frequency information into DL frameworks.
Moreover, different sub-bands contain certain prior information about the image context.
For example, high-frequency sub-bands usually convey more textural details while low-frequency sub-bands deliver more structural information, which can be a strong guide for the network to recover the missing details and maintain the objective quality for image SR~\cite{guo2017deep,huang2017wavelet,deng2019wavelet}.
These ways to introduce the frequency-based transformation into DNNs can be achieved by using the transformed data as the input and output directly~\cite{bae2017beyond,huang2017wavelet,liu2020wavelet,deng2019wavelet,guo2017deep,chen2021all}, as shown
in Fig.~\ref{fig:18-FreqPrior}(a) or applying transformation to the intermediate feature map~\cite{liu2018multi,rong2020burst,Fu_2021_CVPR}, as shown in Fig.~\ref{fig:18-FreqPrior}(b).

In addition, priors using frequency-based transformations can also be encapsulated as powerful plug-and-play modules for the existing DL pipelines.
MWCNN \cite{liu2018multi} replaces the pooling operation with DWT and uses its inverse transformation as the up-sampling operation. As such, it can enlarge the receptive field of DNN models while capturing both frequency and location information from the feature maps.
Along this line, Rong \etal~\cite{rong2020burst} also proposed a wavelet pooling operation based on the Haar wavelet for better denoising and corresponding signal reconstruction.
Moreover, Maggioni \etal~\cite{maggioni2021efficient} introduced learnable parameters into such invertible transformation operations to further improve the DNN model performance.

\textbf{Discussion:}
From our review, the existing DNN models, \eg,~\cite{liu2020wavelet}, that leverage the frequency-based transformation as prior is mostly based on empirical studies.
However, strong task-specific knowledge and extensive experiments are required.
Therefore, it is beneficial to conduct more theoretical analysis, \eg,~\cite{bae2017beyond}, and explore a more general prior using the frequency-based transformation for different deep image restoration and enhancement tasks.
Furthermore, existing works~\cite{liu2018multi,rong2020burst} mostly focus on replacing the pooling operation with priors using the aforementioned transformations. In a nutshell, future research could consider designing more flexible frequency-based transformation modules by introducing learnable parameters, such as ~\cite{maggioni2021efficient}.

\subsubsection{Learning with Non-frequency-based Transformation}
There also exist some non-frequency-based transformations that can serve as informative prior for deep image restoration and enhancement by emphasizing some significant patterns of images.
Among them, edge information obtained from the image can be regarded as informative prior because it can provide detailed structural or geometric guidance for DNN training.

Yang \etal~\cite{yang2017deep} proposed a representative work taking the low-resolution (LR) canny edge map and LR image as inputs to predict the high-resolution (HR) image and HR edge map, respectively. This forces the network to preserve the structural information in SR. Nazeri \etal~\cite{nazeri2019edge} proposed a two-stage SR framework that first predicts an HR edge map from the concatenation of the LR canny edge map and LR image. They then designed an image completion module to obtain an HR image by taking the predicted HR edge map and LR image as inputs.
To better extract edge information, Ren \etal~\cite{ren2020single} and Zhu \etal~\cite{zhu2020eemefn} applied the deep edge detectors, \eg, HDE~\cite{xie2015holistically} to generate the edge maps instead of using the hand-craft operators.
Moreover, Fang \etal~\cite{fang2020soft} proposed a soft-edge assisted framework to jointly learn the edge maps along with the SR task in an end-to-end manner by optimizing the edge loss and SR loss simultaneously.

Beyond edge transformation, Ren \etal~\cite{ren2018gated} also explored using the transformations, including white balance, contrast enhancing, and gamma correction as prior, for deep image dehazing.
In particular, they proposed a gated fusion network taking all the transformed hazy images as inputs to predict their corresponding confidence maps. These confidence maps are then weighted with the transformed images to obtain a clean image.

\textbf{Discussion:}
Applying the selected transformation to the degraded images as prior has shown effectiveness for deep image restoration and enhancement tasks.
However, it would be more meaningful if we could introduce the prior based on the transformation into network design by replacing the hand-crafted transformation operators with a joint sub-network \cite{fang2020soft}.

\subsection{Temporal Prior}
\label{subsec:temporal_prior}

\noindent \textbf{Insight:} \emph{Temporal prior is tailored for deep video restoration and enhancement tasks. This prior in the video is mainly derived from the relationship between frames, \ie, the temporal information.}

In this section, we introduce two main temporal priors: optical flow and temporal sharpness prior. They are commonly used to align the temporal features in deep video restoration and enhancement tasks.
\subsubsection{Optical Flow}
\label{subsubsec:optical_flow}
Optical flow reflects the motion of objects, surfaces, and edges between consecutive frames of a video, caused by the observer and scene~\cite{brox2010large}. It can be formulated as follow:
\begin{equation}
    \small
    I(x,y,t)=I(x+\Delta x,y+\Delta y,z+\Delta z)
    \label{eq:optical_flow},
\end{equation}
where ($x,y$) is the pixel location in an image $I$, $t$ is the time variation between two frames, calculated by the optical flow. Traditional methods focus on the sparse optical flow, which only estimates the pixels with distinct features in the image, \eg, variational approaches~\cite{brox2010large}, Lucas–Kanade method~\cite{bruhn2005lucas}. However, most DL-based methods focus on dense optical flow, which can alleviate the limitations of traditional methods. Dosovitskiy \etal~\cite{dosovitskiy2015flownet} introduced FlowNet, the first CNN-based method for optical flow estimation. This baseline has been continuously improved by, \eg, ~\cite{ilg2017flownet,ranjan2017optical,sun2018pwc},  which have been widely used in various deep image restoration and enhancement tasks.

\begin{figure}
    \centering
    \includegraphics[width=0.49\textwidth]{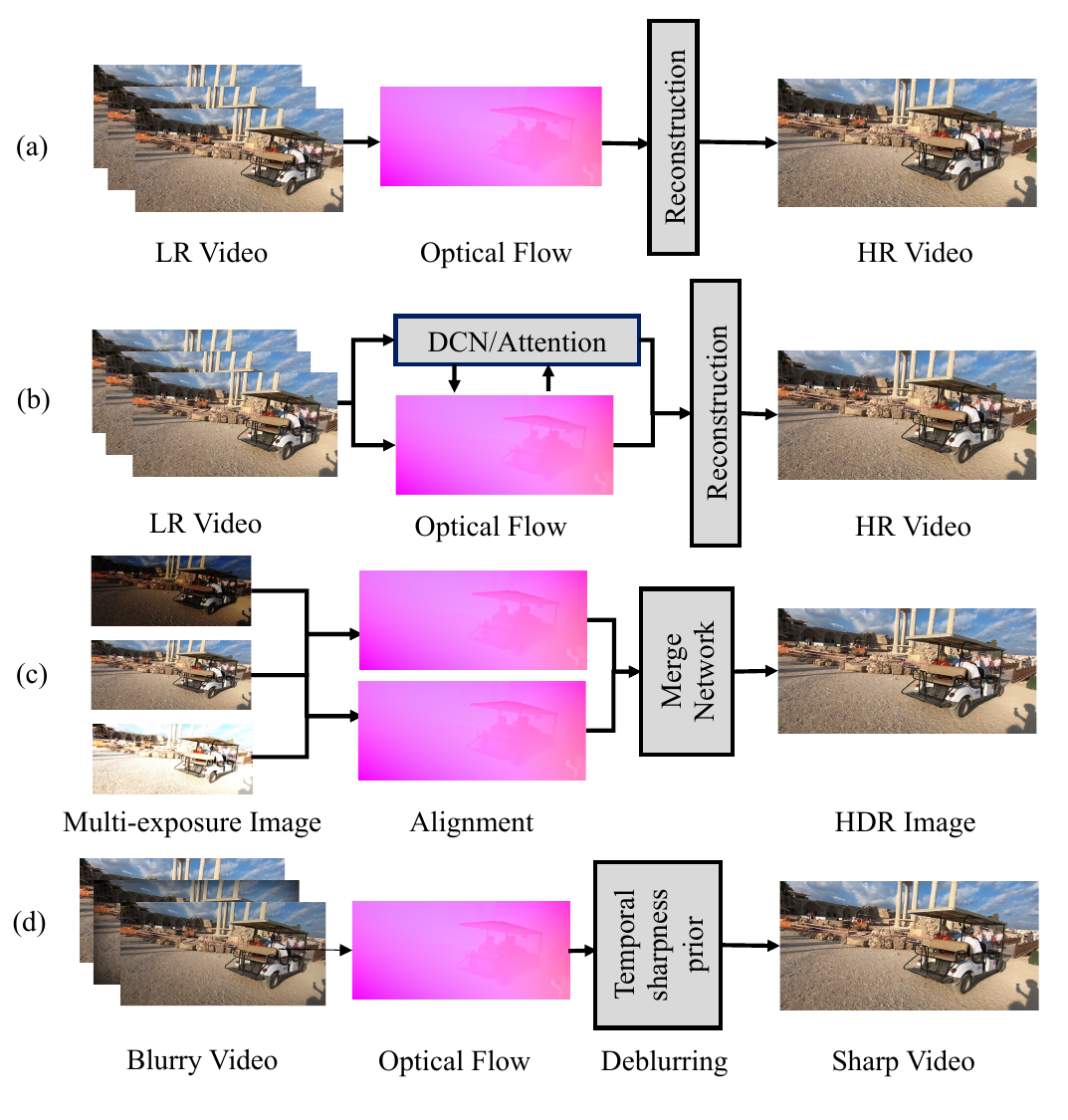}
    \caption{Different usages of optical flow as prior. (a) Directly using the optical flow to align temporal features in VSR. (b) Combining optical flow with DCN or attention to align temporal features in VSR. (c) Merging the multi-exposure images using optical flow in HDR imaging. (d) Constructing the temporal sharpness prior to optical flow in video deblurring.}
    \label{fig:15-VSR-1}
    \vspace{-8pt}
\end{figure}

Firstly, video super-resolution (VSR) aims to recover high-resolution (HR) frames via motion estimation from neighboring frames for the frame alignment. Consequently, optical flow is often used to align the neighboring frames, as shown in Fig.~\ref{fig:15-VSR-1}(a).
Sajjadi \etal~\cite{sajjadi2018frame} proposed a representative framework that relies on a flow-based encoder-decoder architecture to estimate the optical flow.
They also designed a recurrent network to propagate the long-term temporal information to recover HR videos.
ToFlow~\cite{xue2019video} jointly trains the flow estimation component and video processing component to better calculate optical flow for VSR.
However, optical flow is restricted by the invariance of light and large motion. Therefore, several works, \eg, DUF~\cite{jo2018deep}, DCN~\cite{wang2019edvr,tian2020tdan}, have tried to use the CNN-based alignment methods to achieve the frame alignment.
In addition, many recent studies use optical flow as the auxiliary information to support the frame alignment, as shown in Fig.~\ref{fig:15-VSR-1}(b). BasicVSR++~\cite{chan2021basicvsr++} is a representative framework that uses optical flow to guide the DCN for the frame alignment. This work won three champions and one runner-up in NTIRE 2021 Video Restoration and Enhancement Challenge~\cite{son2021ntire}. Similar to BasicVSR++, Liang \etal~\cite{liang2022vrt} used optical flow to guide the self-attention structure to accomplish the temporal feature alignment.

Optical flow is also utilized to deal with the inconsecutive blurred frames in video deblurring, as shown in Fig.~\ref{fig:15-VSR-1}(c).
Pan \etal~\cite{pan2020cascaded} proposed the first work to use the warped adjacent frames obtained by optical flow to construct the temporal sharpness prior, which will be discussed in the following section.
Some recent works \cite{li2021arvo,shang2021bringing} also adopt a similar idea by taking the optical flow as a prerequisite for obtaining the temporal sharpness prior.
Different from works \cite{pan2020cascaded,li2021arvo,shang2021bringing}, Wang \etal~\cite{wang2022efficient} propose using optical flow to estimate motion magnitude prior during image caption as blur degree.
This method incorporates optical flow as a prior for blur estimation, thereby effectively directing the deblurring network to focus on distinct regions.

Moreover, optical flow is used to align the low-, medium-, and high-exposure frames in multi-exposure HDR imaging, as shown in Fig.~\ref{fig:15-VSR-1}(d). Representative works use classic optical flow estimation methods~\cite{kalantari2017deep,yan2019multi} or DL-based methods~\cite{prabhakar2019fast,peng2018deep} to align the multi-exposure LDR images. They then design DNN-based frameworks to fuse the aligned LDR images to reconstruct the HDR images.
Tab. \ref{table1:optical_flow} summarizes  representative tasks using optical flow as prior for deep image/video restoration and enhancement. In addition to the approaches shown in Fig.~\ref{fig:15-VSR-1}, optical flow is commonly used in other tasks,~\eg, deep video deraining\cite{yang2019frame}, deep video frame interpolation \cite{bao2019memc}.

\begin{table*}[t!]
\renewcommand{\tabcolsep}{3pt}
\caption{Applications of optical flow as prior for deep video restoration and enhancement tasks}
\vspace{-8pt}
\label{table1:optical_flow}
\begin{center}
\resizebox{\linewidth}{!}{
\begin{tabular}{ccccc}
\hline
    Method &Publication&Task&Optical flow approach& Highlight\\
\hline
\hline
    Sajjadi \cite{sajjadi2018frame} &CVPR 2018& VSR & FNet  &  Recurrent framework with optical flow estimation network(FNet)\\
    Xue \cite{xue2019video}  &IJCV 2019& VSR & SPyNet\cite{ranjan2017optical} & Task-oriented optical flow framework \\
    Chan \cite{chan2021basicvsr++} &NTIRE 2021& VSR & SPyNet\cite{ranjan2017optical} & Use optical flow to guide the DCN\\
    Pan \cite{pan2020cascaded} & CVPR 2020&Video deblurring & PWC-Net\cite{sun2018pwc} & Use optical flow to construct the temporal sharpness prior\\
    Li \cite{li2021arvo} &CVPR 2021& Video deblurring & PWC-Net\cite{sun2018pwc} & Use optical flow to construct the temporal sharpness prior\\
    Shang \cite{shang2021bringing} &CVPR 2021& Video deblurring & PWC-Net\cite{sun2018pwc} & Use optical flow to construct the temporal sharpness prior\\
    Wang \cite{wang2022efficient} & ECCV 2022 & Video deblurring    & RFAT \cite{teed2020raft} & Use optical flow to estimate motion magnitude prior\\
    Yang \cite{yang2019frame} &CVPR 2019& Video deraining &FlowNet\cite{dosovitskiy2015flownet}& Optical flow help extract the temporal rain-related feature\\
    Prabhakar \cite{prabhakar2019fast} &ICCP 2019& Multi-Exposure HDR &PWC-Net\cite{sun2018pwc}& Align multi-exposure image\\
    Bao \cite{bao2019memc} &TPAMI 2019&Video frame interpolation &PWC-Net\cite{sun2018pwc}& Lead the flow-based motion interpolation algorithms\\
\hline
\end{tabular}
}
\end{center}
\vspace{-14pt}
\end{table*}

\textbf{Discussion}: From the aforementioned analysis, it is difficult to calculate optical flow in case of the variational lighting environment and large motion. Consequently, these problems are inevitable in real-world videos.
Although many works \cite{jo2018deep,wang2019edvr,tian2020tdan} try to replace optical flow with DNNs for the frame alignment, important motion information contained in optical flow is lost.
For this reason, active research has been conducted to alleviate the limitations for better frame alignment. Many recent works have shown that using optical flow estimation as a preliminary or auxiliary component combined with other methods is better than those using priors only for deep video restoration and enhancement.
For example, \cite{chan2021basicvsr++,liang2022vrt} use optical flow to guide the DCN or self-attention for VSR, and \cite{pan2020cascaded,li2021arvo,shang2021bringing} use the optical flow to generate the temporal sharpness prior for video deblurring.
Future research could consider combining optical flow with DNNs to tackle the challenges for real-world applications.

\subsubsection{Temporal Sharpness Prior}
\label{subsubsec:temporal_harpness_prior}

Temporal Sharpness Prior is a specific prior in deep video deblurring based on the hypothesis of nonconsecutive blur property~\cite{pan2020cascaded}. The same object in different frames contains sharp pixels and blurred pixels simultaneously. Sharp pixels can be priors for deep video deblurring. Although some traditional methods~\cite{cho2012video} have used this prior, it is firstly combined with a DL-based model and explicitly defined in ~\cite{pan2020cascaded}. Pan \etal~\cite{pan2020cascaded} proposed a representative framework to compare the current frame with its warped one, which is obtained using the optical flow from adjacent frames to construct the temporal sharpness prior. This can constrain the deep CNN-based models to exploit the sharp frame to support deep video deblurring. A similar approach was proposed by Li \etal~\cite{li2021arvo} to deal with large motion in the video; however, it takes multiple consecutive frames as inputs to calculate the temporal sharpness prior to acquire more decisive temporal information.
Moreover, its feature space can be matched on multiple scales. Shang \etal~\cite{shang2021bringing} proposed a bidirectional long short-term memory (LSTM) architecture to probe the two nearest sharp frames (NSFs) from the front and rear neighboring frames. The reason is that the temporal sharpness prior obtained from the neighboring frames within a reasonable range can effectively facilitate deep video deblurring.

\textbf{Discussion}: From our review, temporal sharpness prior is mostly based on the nonconsecutive blur property in the consecutive frames~\cite{pan2020cascaded,li2021arvo,shang2021bringing}.  This concept can be easily extended to other domains. For example, because noise is also nonconsecutive in the video, one can use the same assumption to construct another prior. Therefore, future research could focus more on specific temporal properties to explore new temporal prior for deep video restoration and enhancement.

\subsection{Non-local Self-similarity as Prior}
\label{sec:non}
\begin{figure}[t]
\centering
\includegraphics[width=0.48\textwidth]{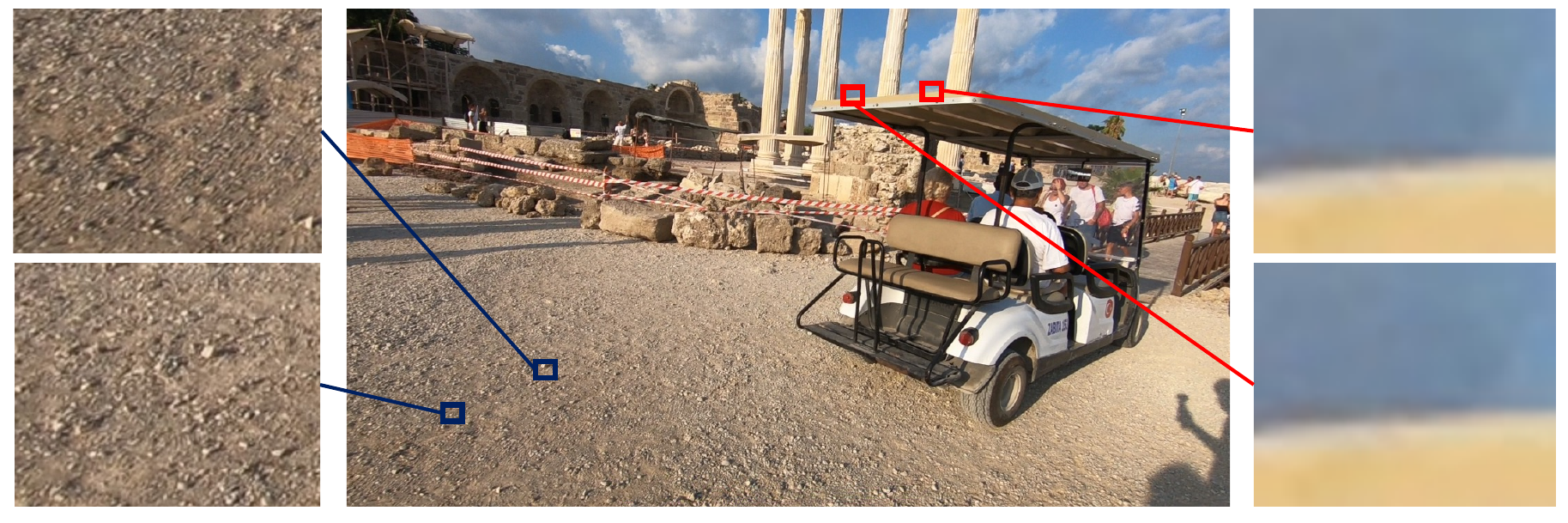}
\caption{Illustration of small patterns that tend to reappear in the image.}
\label{10-sim}
\centering
\vspace{-12pt}
\end{figure}

\noindent \textbf{Insight:} \textit{Non-local self-similarity prior can help restore and enhance specific details with the reappearance patches in an image.}

Non-local self-similarity is a fundamental property of an image, which indicates that the same small patterns tend to reappear in a given image\cite{dabov2007image}, as shown in Fig. \ref{10-sim}. Therefore, many low-level vision tasks utilize this propriety as a prior to facilitate the recovery of high-quality images.
Traditional methods,~\eg, 3D transform BM3D\cite{dabov2007image}, WNNM\cite{gu2014weighted} use mathematical transformation and mapping to stack similar patches as prior to achieve the restoration task. However, these methods heavily rely on hand-crafted transformations to learn the self-similarity.

DL-based methods, on the other hand, combine DNNs with the non-local self-similarity prior for tasks, such as denoising\cite{lefkimmiatis2017non}, SR\cite{mei2020image,mei2021image}, and image restoration\cite{zhang2019residual}. Lefkimmiatis \etal~\cite{lefkimmiatis2017non} proposed the first DL-based framework to group similar patches and jointly filter them. However, the prior itself is heavily dependent on the manual design, and the method shows limited generalization capacity.
By contrast, \cite{zhang2019residual} proposed a representative framework that tackles the problems in \cite{lefkimmiatis2017non} via a non-local attention network~\cite{wang2018non}. The reason is that non-local attention conforms well to the properties of the non-local self-similarity prior, making it more effective for DL-based models and relieving the need for manual design.
Mei \etal~\cite{mei2021image} extended this idea to non-local sparse attention to capture the long-range similarity, \ie, global similarity, information.

\textbf{Discussion}: From the above analysis, we assume that non-local attention~\cite{wang2018non} can effectively replace the role of the manually designed non-local self-similarity prior. However, as pointed out by \cite{mei2021image}, non-local attention suffers from two problems: limited attention to local features and high computational costs.
Therefore, it deserves exploring the sparse attention or other attention frameworks to better extract the global similarity features for deep image restoration and enhancement.

\section{Statistical Priors}
\label{primary-prior:statistic}

Statistical priors are based on the statistical features and noise models in imaging, providing feature distribution about the image content.
We divide the statistical priors into three parts: {image intensity and gradient distribution}, {physics-based prior}, and {kernel and noise information as prior}.

Image intensity and gradient distribution priors (Sec. \ref{sec:statistical}) utilize statistical properties of the image content.
For example, the dark channel prior \cite{pan2016blind} and the bright channel prior \cite{cai2020dark} are widely used for parameter estimation in deblurring tasks.
Physics-based priors (Sec. \ref{sec:physical}) incorporate physical models and principles into the restoration and enhancement of the image.
For example, the atmospheric scattering model \cite{narasimhan2003contrast} and rain model \cite{li2016rain} utilize the physical properties of the atmosphere and rain to guide dehazing and deraining tasks.
These priors can provide statistical constraints and guidance by incorporating physical models and principles, improving image quality.
Kernel and noise information prior (Sec. \ref{subsec:kernel_prior}) is another statistical prior that leverages the knowledge of the kernel and noise models of the image to guide the deblurring or denoising~\cite{zhang2018learning,gu2019blind}.
By incorporating the statistical distribution of kernel and noise, these priors can help estimate blur kernel and reduce noise.

\begin{figure}[t!]
    \centering
    \includegraphics[width=0.47\textwidth]{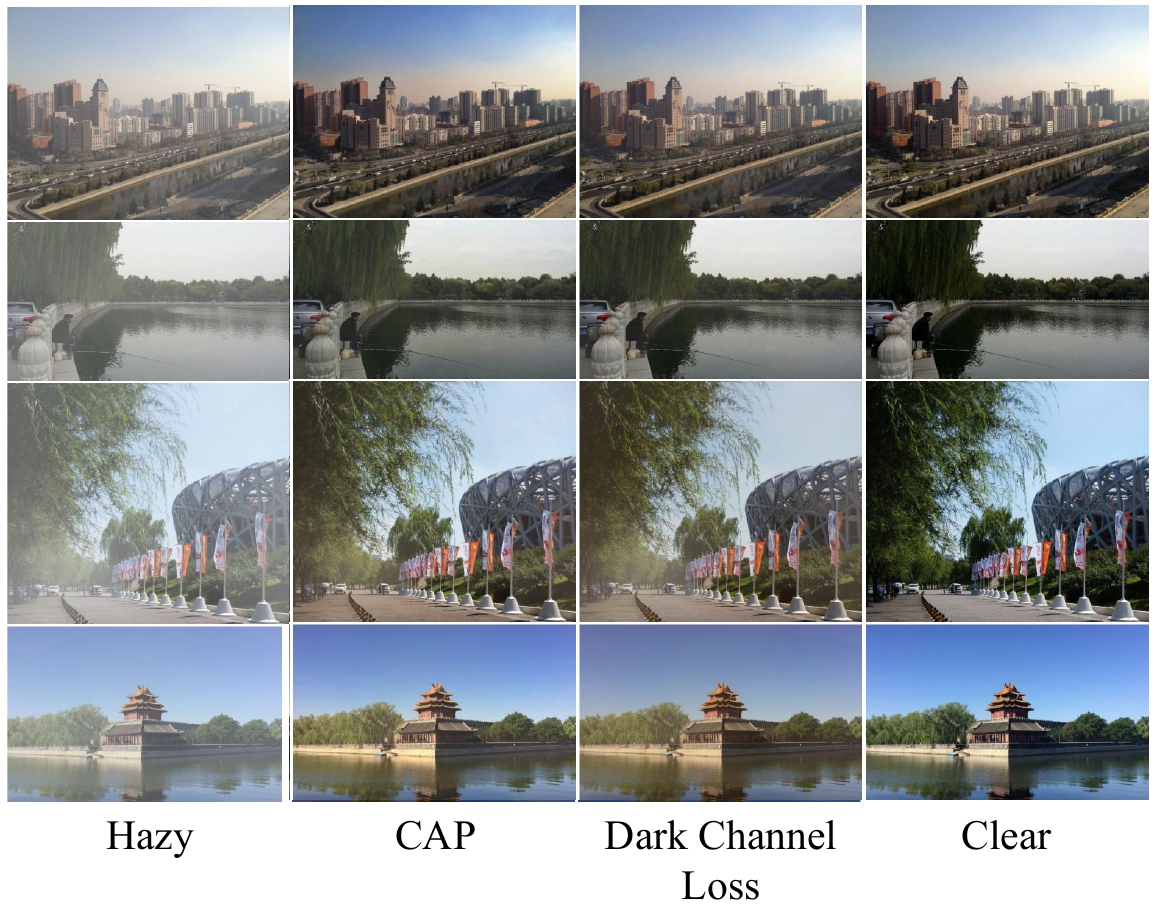}
    \caption{Qualitative results on RESIDE’s HSTS.
    CAP\cite{zhu2015fast}, With Dark Channel Loss\cite{golts2019unsupervised}.}
    \label{fig:A-202-Dehazing-w_wo_dcploss}
    \vspace{-8pt}
\end{figure}

\vspace{-20pt}
\subsection{Image Intensity and Gradient Distribution as Prior}
\label{sec:statistical}

\noindent \textbf{Insight:} \textit{Image intensity and gradient distribution as prior focus on image $I(x)$ and image derivative $I'(x)$.
}

In this section, we review the representative works that exploit the statistical image feature prior to estimating the parameters of DNNs or embedding them into the DNN structures.
The statistical image feature as prior can be divided into: the statistical intensity feature as prior and the statistical gradient feature as prior.
The representative works in using these priors are listed in Tab. \ref{tab:statistical_feature_table_1}.

\subsubsection{Statistical Intensity Feature as Prior}
\label{subsubsec:intensity_statistical_feature}

Statistical intensity features usually refer to the image channel intensity distribution, \eg, two-tone distribution prior \cite{pan2016l_0}, two-color prior \cite{joshi2009image}, histogram equalization prior \cite{dhal2021histogram,fu2020underwater}, and extreme value distribution, \eg, dark channel prior \cite{pan2016blind,he2010single}, bright channel prior \cite{yan2017image,tao2017low,lee2020unsupervised}.

\begin{figure}[t!]
    \centering
    \includegraphics[width=0.5\textwidth]{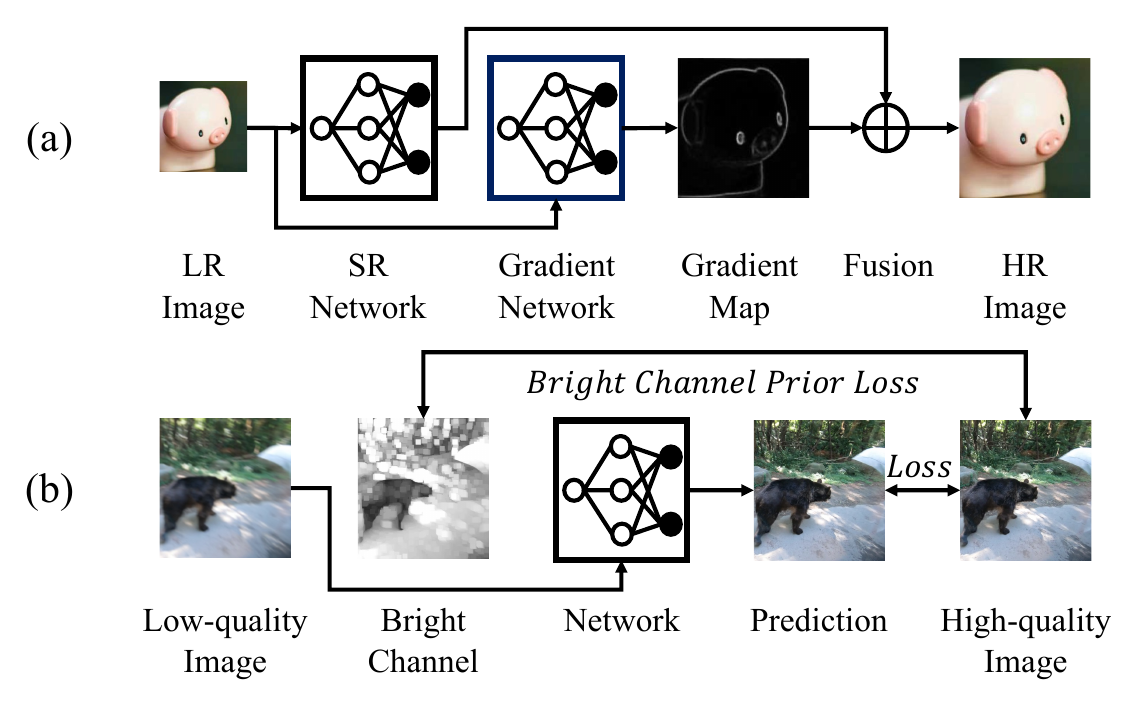}
    \caption{Three representative frameworks use statistical features as prior.
    (a) exploits a deep network to predict gradient maps of images to guide the recovery of high-resolution details \cite{ma2020structure}.
    (b) employs bright channel prior as a loss function to support an unsupervised learning approach for single low-light image enhancement \cite{lee2020unsupervised}.
    }
    \label{fig:101}
    \vspace{-10pt}
\end{figure}

\begin{figure}[t]
    \centering
    \includegraphics[width=0.5\textwidth]{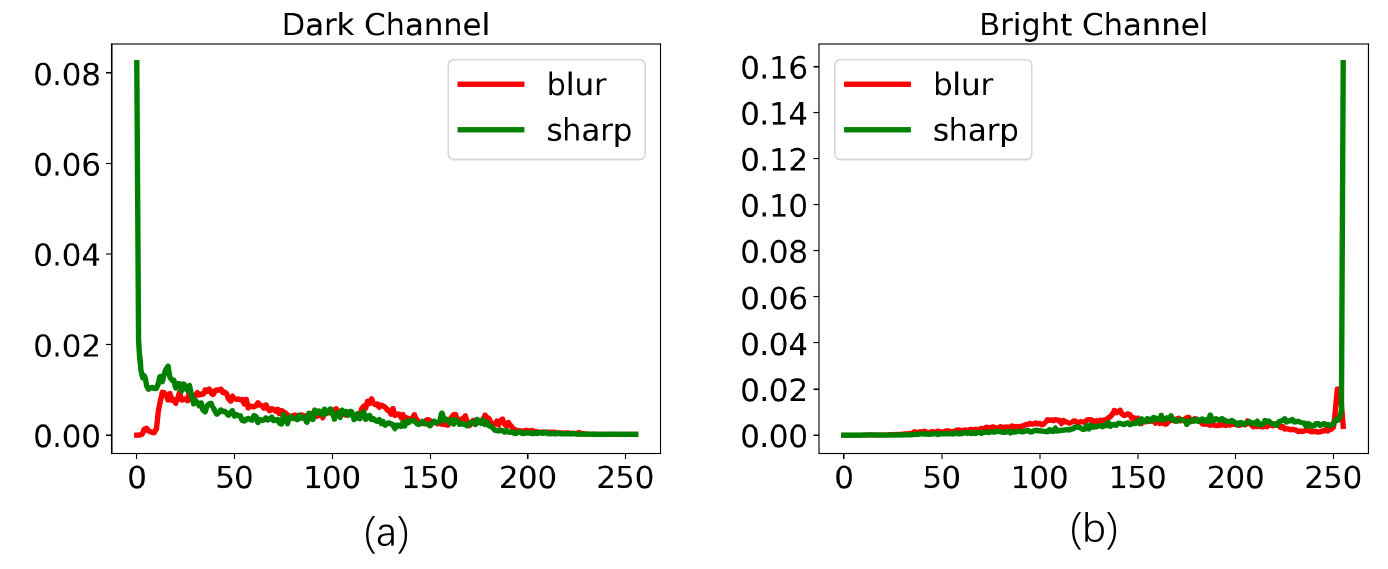}
    \caption{(a) and (b) represent the statistics of the dark and bright channel priors from  the blur images and sharp images.}
    \label{fig:13-dark-channel}
\end{figure}

Dark channel prior \cite{he2010single} describes that at least one of the color channels in a local neighborhood possesses pixels with intensity values close to zero.
Yan \etal~\cite{yan2017image} extended it to the bright channel prior, which constrains that at least one of the color channels in small neighbor pixels has a very large intensity, as shown in Fig. \ref{fig:13-dark-channel}.
They are widely used for kernel estimation or parameter constraints in traditional image restoration tasks, \eg, dehazing, denoising, and low-light enhancement.
Lee \etal~\cite{lee2020unsupervised} proposed the first unsupervised framework leveraging the bright channel prior for the low-light image enhancement. This prior predicts the initial illumination map in the image, which is then used to build up the unsupervised loss to optimize the encoder-decoder network.
By contrast, ECPeNet\cite{cai2020dark} directly plugs the dark channel prior and bright channel prior into the encoder-decoder network for image deblurring. As such, this approach can aggregate information of both priors and blurry image representations to guide the dynamic scene deblurring.

\begin{table}[t!]
\caption{Application of statistical feature as prior in deep image/video restoration and enhancement tasks.}
\vspace{-10pt}
\label{tab:statistical_feature_table_1}
\begin{center}
\setlength{\tabcolsep}{1.2mm}{
\resizebox{\linewidth}{!}{
    \begin{tabular}{cccc}
\hline
    Method                         & Task                                      & \makecell[c]{Statistical Feature}          & Highlight \\
\hline
\hline
    TIP 2018 \cite{golts2019unsupervised}    & Dehazing                  & Dark channel prior                    & Loss function \\
    ISPL 2020\cite{lee2020unsupervised} & \makecell[c]{Low-light}     & Bright channel prior                & \makecell[c]{Unsupervised \\ learning} \\
    TIP 2020\cite{cai2020dark}       & Deblurring                                & \makecell[c]{Dark and bright \\ channel prior}  & \makecell[c]{Combine Dark and \\ bright channel prior } \\
    CVPR 2020\cite{ma2020structure}     & \makecell[c]{SR}          & Gradient guidance             & \makecell[c]{Gradient maps} \\
    CVPR 2021\cite{sun2021learning}     & \makecell[c]{SR}          & Gradient guidance             & \makecell[c]{Structure guidance by \\ gradient prior} \\
\hline
\end{tabular}}}
\end{center}
\vspace{-8pt}
\end{table}

\subsubsection{Statistical Gradient Feature as Prior}
\label{subsubsec:gradient_statistical_features}

Statistical gradient feature refers to the image gradient distribution and extreme value distribution of the local gradient, including local maximum gradient prior \cite{chen2019blind}, gradient guidance prior \cite{ma2020structure}, gradient channel prior \cite{singh2019single,kaur2020color}.
Despite their widespread applications in traditional image restoration and enhancement tasks, they have rarely been explored for DL-based frameworks.

Ma \etal~\cite{ma2020structure} proposed a gradient branch network using the gradient as a prior to obtain additional structural information and construct a gradient loss. This prior enables the network to concentrate more on the geometric structures for image SR, as shown in Fig. \ref{fig:101}(a).
Sun \etal~\cite{sun2021learning} followed a similar idea to attain the structural information to guide the SR network.

\textbf{Discussion:}
From the review, we identify that the prior based on the statistical gradient feature can provide extra knowledge or guidance for deep image restoration and enhancement tasks~\cite{ma2020structure}.
However, some of the aforementioned priors, \eg, local maximum gradient prior~\cite{chen2019blind}, gradient channel prior~\cite{singh2019single,kaur2020color}, have not yet been applied to the DL-based methods for deep image restoration and enhancement.
Therefore, future research could possibly focus on this direction.

\subsection{Physics-based Prior}
\label{sec:physical}
\noindent \textbf{Insight:}\textit{
Physical models can reduce the learning difficulty and increase interpretability by decomposing the output into separate components or factors following the physical principles.
}

\begin{table*}[t!]
\centering
\caption{Representative deep image restoration and enhancement works with physics-based prior}
\label{tab:physical_based_prior}
\resizebox{\linewidth}{!}{
\begin{tabular}{ccccc}
\hline
 Method &Publication &  Task & Physics-based Prior & Highlight \\
\hline
\hline
Cai~\cite{cai2016dehazenet}  &TIP 2016 &  Dehazing & ASM &
Estimate transmission map\\
Li~\cite{li2017aod} &ICCV 2017 &  Dehazing & ASM & Estimate an intermediate parameter by reformulated ASM \\
Zhang~\cite{zhang2018densely} &CVPR 2018 &  Dehazing & ASM & Jointly estimate transmission map and atmospheric light\\
Zhang~\cite{zhang2019famed} &TIP 2019 &  Dehazing & ASM & Estimate an intermediate parameter by reformulated ASM \\
Dong~\cite{dong2020physics} &ECCV 2020 &  Dehazing & ASM & Jointly estimate transmission map and atmospheric light\\
Chen~\cite{chen2020jstasr} &ECCV 2020 &  Desnowing & ASM & Jointly estimate transmission map and atmospheric light \\
Kar~\cite{kar2021zero} &CVPR 2021 &  Dehazing;Low-light & ASM &  Jointly estimate transmission map and atmospheric light \\

\hline
Fu~\cite{fu2017removing} &CVPR 2017 &  Deraining & Rain Model & Residual learning \\
Yang~\cite{yang2017deep2}  &CVPR 2017 &  Deraining & Rain Model & Recurrent Residual learning with multiple steak layer and rain mask\\
Li~\cite{li2018recurrent}  &ECCV 2018 &  Deraining & Rain Model & Recurrent Residual learning with multiple steak layer\\
Liu~\cite{liu2018erase}  &CVPR 2018 &  Video Deraining & Rain Model & Residual learning considering occlusion\\
Hu~\cite{hu2019depth}  &CVPR 2019 &  Deraining & Rain Model & Residual learning with depth information guidance \\
Yang~\cite{yang2019frame}  &CVPR 2019 &  Video Deraining & Rain Model & Residual learning with temporal fusion \\

\hline
Chen~\cite{DBLP:conf/bmvc/WeiWY018} & BMVC 2018 &  Low-light & Retinex Model & Estimate reflectance and illumination \\
Li~\cite{li2018lightennet} &PRL 2018 &   Low-light & Retinex Model & Estimate illumination\\
Zhang~\cite{zhang2019kindling} & ACM MM 2019 &  Low-light & Retinex Model & Estimate reflectance and illumination \\
Zhang~\cite{zhang2021beyond} &IJCV 2021 &   Low-light & Retinex Model & Estimate reflectance and illumination \\
Yang~\cite{yang2021sparse} &TIP 2021 &   Low-light & Retinex Model & Estimate reflectance and illumination\\
Liu~\cite{liu2021retinex} &CVPR 2021 &   Low-light & Retinex Model & Estimate illumination\\
\hline
\end{tabular}
}
\vspace{-10pt}
\end{table*}

\begin{figure}[t!]
    \centering
    \includegraphics[width=0.48\textwidth]{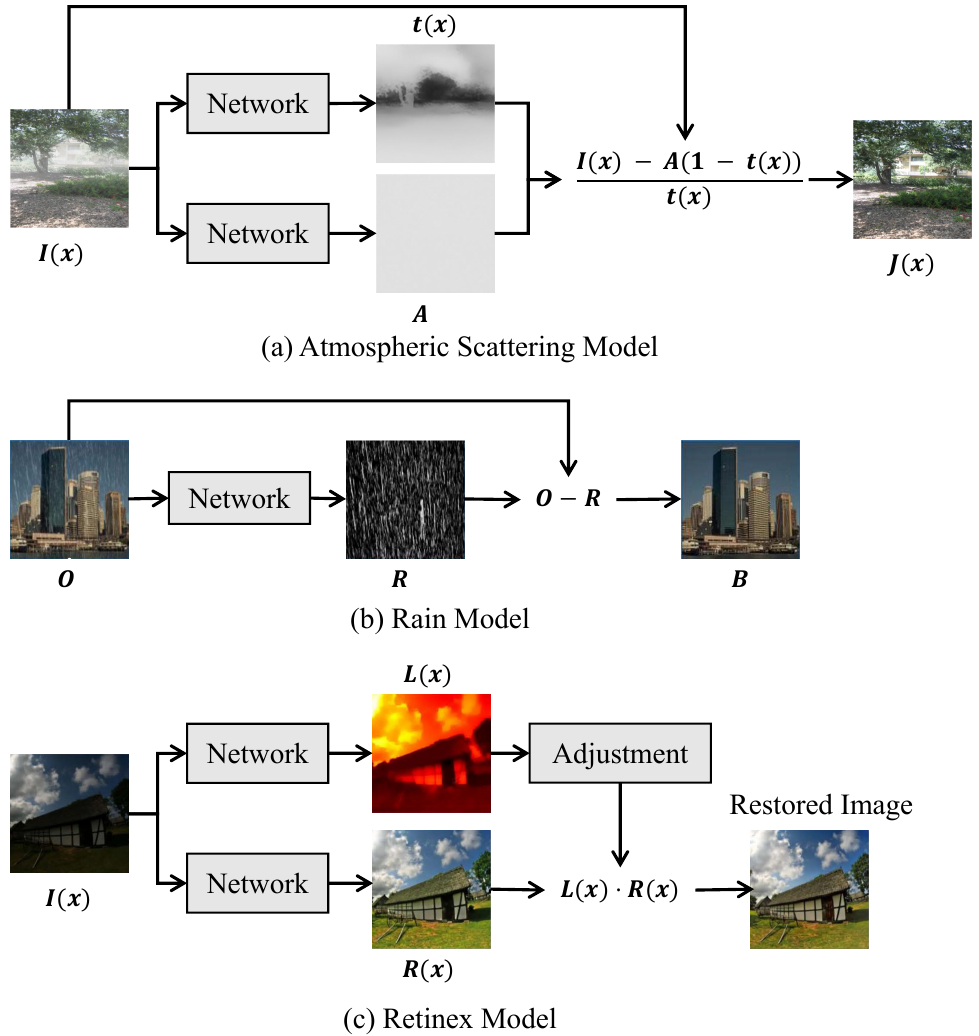}
    \caption{Representative DNN frameworks using physics-based prior for image restoration and enhancement tasks.
    (a)-(c) are simplified illustrations from ~\cite{zhang2018densely},
    ~\cite{fu2017removing}, ~\cite{zhang2019kindling}, respectively
    }
    \label{fig:physical}
    \vspace{-12pt}
\end{figure}

Bad weather, \eg, haze, rain, and snow, causes image quality degradation during the image capturing process.
Traditional methods, \eg,~\cite{narasimhan2003contrast} have carefully designed degradation models following the physical principles.
Although many works directly restore a clean image from the degraded one using a DNN, some attempts have been made by introducing the traditional methods into the DL pipeline as a strong prior.
Guided by the prior knowledge within these physical models, as shown in Fig.~\ref{fig:physical}, the designed DNN models are more effective and interpretable, stimulating the development of deep image restoration and enhancement methods.

\subsubsection{Atmospheric Scattering Model}
\label{subsubsec:atmospheric_scattering_model}
One representative degradation model is the atmospheric scattering model (ASM)~\cite{narasimhan2003contrast}, which is widely adopted for deep image dehazing~\cite{cai2016dehazenet}.
The formula of ASM is $I(x) = J(x)t(x) + A(1 - t(x))$,
where $x$ represents the position of pixel, $t(x)$ is the medium transmission map and $A$ stands for the global atmospheric light. $J(x)$ is the clean image, and $I(x)$ denotes the degraded image in a hazy scene.
Following the ASM model, DehazeNet~\cite{cai2016dehazenet} uses a CNN to estimate transmission map $t(x)$ and obtain atmospheric light $A$ from 0.1\% darkest pixels of the estimated $t(x)$.
To obtain a more accurate atmospheric light $A$, DCPDN~\cite{zhang2018densely} proposes to jointly learn a densely connected pyramid network to estimate the transmission map $t(x)$ and the UNet~\cite{ronneberger2015u} to estimate atmospheric light $A$.
Moreover, to reduce the cumulative error caused by the separately estimated $t(x)$ and $A$, many methods have tried to use a single intermediate parameter by reformulating the ASM model.
For instance, in AOD-Net~\cite{li2017aod}, ASM model is reformulated by an intermediate parameter $K(x)$. Thus ASM model can be solved by $J(x) = K(x)I(x) - K(x) + b$.
The promising performance shown by AOD-Net~\cite{li2017aod} proves that the ASM model can still be embedded into the DL framework well without the separate estimation of $t(x)$ and $A$.
Later on, DehazeGAN~\cite{zhu2018dehazegan}, FAMEDNet~\cite{zhang2019famed}, and PFDN~\cite{dong2020physics} also follow such a simplified ASM model.
Moreover, the ASM model can be used in other tasks, such as desnowing~\cite{chen2020jstasr} and low-light image enhancement~\cite{kar2021zero} to guide the DNN model design by estimating the corresponding components.

\subsubsection{Rain Model}

Various rain models have been proposed to better adapt different rain patterns for deep image deraining.
Additive composite model (ACM)~\cite{li2016rain} is the simplest and most widely used rain model, which can be formulated as $O(x) = B(x) + S(x)$,
where $x$ represents the position of a pixel, $B$ denotes the clear background layer, $S$ is the rain streak layer and $O$ is the degraded image caused by rain streaks.
Based on ACM, Fu \etal~\cite{fu2017removing} proposed a representative work by
taking the input from the detail layer and
predicting the residual between the degraded image and the ground truth clean image.
Following the idea in \cite{fu2017removing}, DID-MDN~\cite{zhang2018density} introduces an extra rain density classifier to enhance the residual quality.
RESCAN~\cite{li2018recurrent} proposes a simpler recurrent framework to perform multi-stage rain removal via residual learning.
Considering the rain streaks and rain accumulation, JORDER~\cite{yang2017deep2} proposes another representative approach by introducing rain masks and additional rain-streak layers.
Therefore, the degraded rain images in JORDER can be decomposed into multiple rain-streak layers, rain regions, and clean background images as the prediction targets for multi-stage rain removal.
Moreover, some works interpret the rainy scenes with different factors, such as occlusion~\cite{liu2018erase}, temporal properties~\cite{yang2019frame}, or depth map~\cite{hu2019depth}.

\begin{table}[t!]
\centering
\caption{Representative works with modeling kernel and noise information as prior.
A more detailed summary can be found in the supplementary material.}
\label{tab:kernel_based}
\setlength{\tabcolsep}{3mm}{
\resizebox{\linewidth}{!}{
\begin{tabular}{ccc}
\hline
 Method &  Task &  Highlight \\
\hline
\hline
CVPR 2018~\cite{zhang2018learning} &  SR & \makecell[c]{Kernel for  conditional input} \\
TIP 2018 ~\cite{zhang2018ffdnet}  &  Denoising & \makecell[c]{Noise for  conditional input} \\
CVPR 2019~\cite{zhang2019deep} &  SR & \makecell[c]{Kernel for conditional input} \\
NeurIPS 2019~\cite{bell2019blind} &  SR & \makecell[c]{Estimate kernel} \\
CVPR 2019~\cite{gu2019blind}  &  SR & \makecell[c]{Estimate kernel \& \\Kernel for conditional input}\\
CVPR 2019~\cite{guo2019toward} &  Denoising & \makecell[c]{Estimate noise \& \\Noise for conditional input} \\
CVPR 2020~\cite{xu2020unified} &  SR &  \makecell[c]{Kernel for conditional input} \\
NeurIPS 2020~\cite{huang2020unfolding} &  SR & \makecell[c]{Estimate kernel \& \\Kernel for conditional input } \\
CVPR 2021~\cite{liang2021flow}  &  SR &  \makecell[c]{Estimate kernel} \\
\hline
CVPR 2018 ~\cite{chen2018image} &  Denoising & \makecell[c]{Synthetic noise modeling} \\
ICCV 2019~\cite{zhou2019kernel} &  SR & \makecell[c]{Synthetic   kernel modeling}   \\
CVPRW 2020 ~\cite{ji2020real} &  SR & \makecell[c]{Synthetic   kernel modeling} \\
CVPR 2021 ~\cite{tran2021explore} &  Deblur &  \makecell[c]{Synthetic   kernel modeling}  \\
ICCV 2021 ~\cite{jang2021c2n} &  Denoising & \makecell[c]{Synthetic noise modeling} \\
Arxiv 2023 ~\cite{zhang2023kbnet} & \makecell[c]{Denoising \&\\Deblur \&\\Dehazing}  & \makecell[c]{Synthetic kernel modeling} \\
\hline
\end{tabular}}}
\vspace{-8pt}
\end{table}

\subsubsection{Retinex Model}
Motivated by how human perceives color,  the Retinex model \cite{land1986alternative} assumes that an observed image $I$ can be decomposed into two components, reflectance $R$ and illumination $L$.
Formally, the Retinex model can be described as $I(x) = R(x) \cdot L(x)$,
where $x$ denotes the position of a pixel and $\cdot$ means pixel-wise multiplication operation.
Retinex model has been widely integrated into deep low-light image enhancement methods, \eg,~\cite{DBLP:conf/bmvc/WeiWY018,zhang2019kindling}, showing promising performance.
The basic idea of these methods is to design specialized sub-networks for learning the illumination and reflectance components separately.
Retinex-Net~\cite{DBLP:conf/bmvc/WeiWY018} is a representative work that proposes a decomposition network to decompose the input image into reflectance and illumination.
Then, an enhancement network is designed to adjust the illumination map.
Following the pipeline in Retinex-Net, many works~\cite{zhang2019kindling,zhang2021beyond,yang2021sparse} have been proposed with more constraints on the predictions or advanced network architectures.
To improve the computational efficiency, LightenNet~\cite{li2018lightennet} proposes to use a more tiny encoder to estimate illumination and divide the input image by the illumination map to obtain the enhanced image.
Along this line, IM-Net~\cite{wang2019progressive} and RUAS~\cite{liu2021retinex} also merely focus on illumination estimation in DNN framework design.

\begin{figure}
    \centering
    \includegraphics[width=0.48\textwidth]{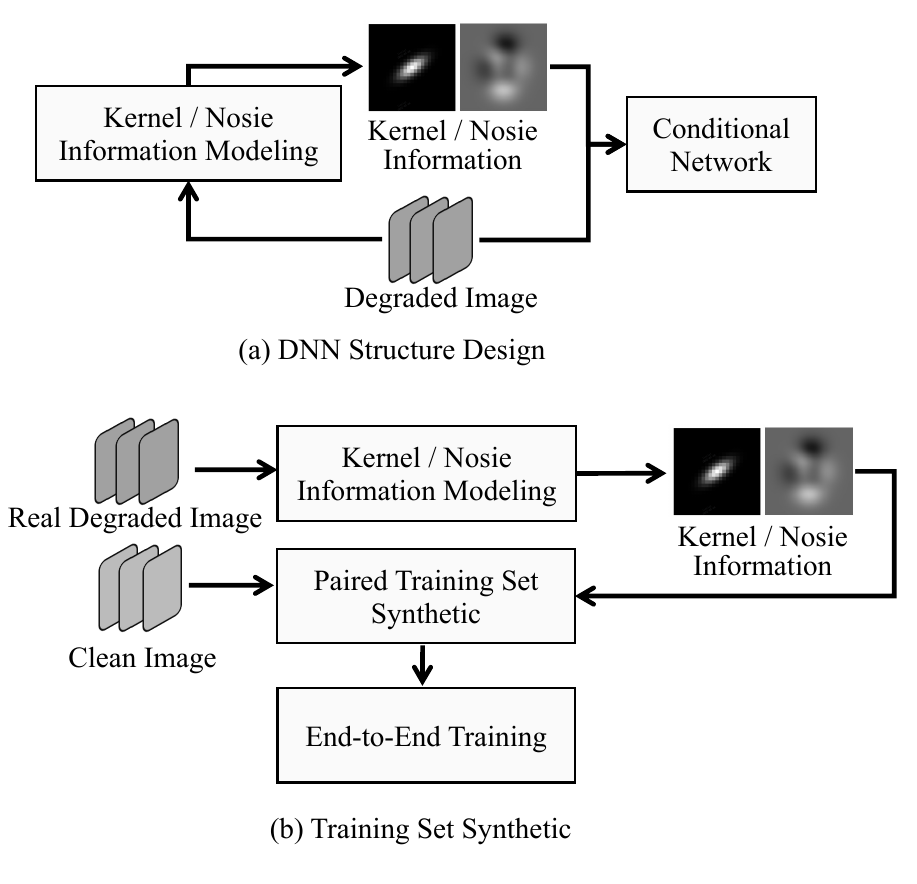}
    \vspace{-10pt}
    \caption{Illustration of pipelines based on modeling kernel and noise information.}
    \label{fig:kernel}
    \vspace{-10pt}
\end{figure}

\textbf{Discussion:} From the review, the DL methods based on the physics-based prior usually rely on the decomposition strategy to predict and assemble different components, which limits their flexibility in DNN framework design.
It is possible to combine the neural architecture search techniques,~\eg,~\cite{li2020all}, with the physics-based prior to spark more effective DNN design schemes.
Moreover, the physical models are usually task-specific or focus on certain application scenarios. Therefore, future research could focus on designing a more general physical model that can be used for various deep image restoration and enhancement tasks.

\vspace{-10pt}
\subsection{Kernel and Noise Information as Prior}
\label{subsec:kernel_prior}
\noindent \textbf{Insight:} \textit{Modeling kernel and noise information can provide extra information for image-specific restoration or help generate more realistic low-quality images.
}

Modeling the image degradation process with pre-defined degradation models can be an effective prior to reduce the solution space.
In the general pipeline for image restoration, a classical degradation model has been widely adopted by many works~\cite{gu2019blind,zhang2018learning}.
The classical degradation model can be formulated as $y = (x \otimes k)\downarrow_s + n \label{eq:degra_model}$,
where $y$ denotes the degraded image, $x \otimes k$ represents the convolution operation between latent image $x$ and kernel $k$, $\downarrow_s$ is the down-sample operation with the scale factor $s$, and $n$ is the noise.
This degradation information, \ie, kernel and noise, can be used as prior in DNN structure design, as shown in Fig.~\ref{fig:kernel} (a) or training dataset synthesis to improve the network performance and robustness, as shown in Fig.~\ref{fig:kernel} (b).

\begin{figure}
    \centering
    \includegraphics[width=0.48\textwidth]{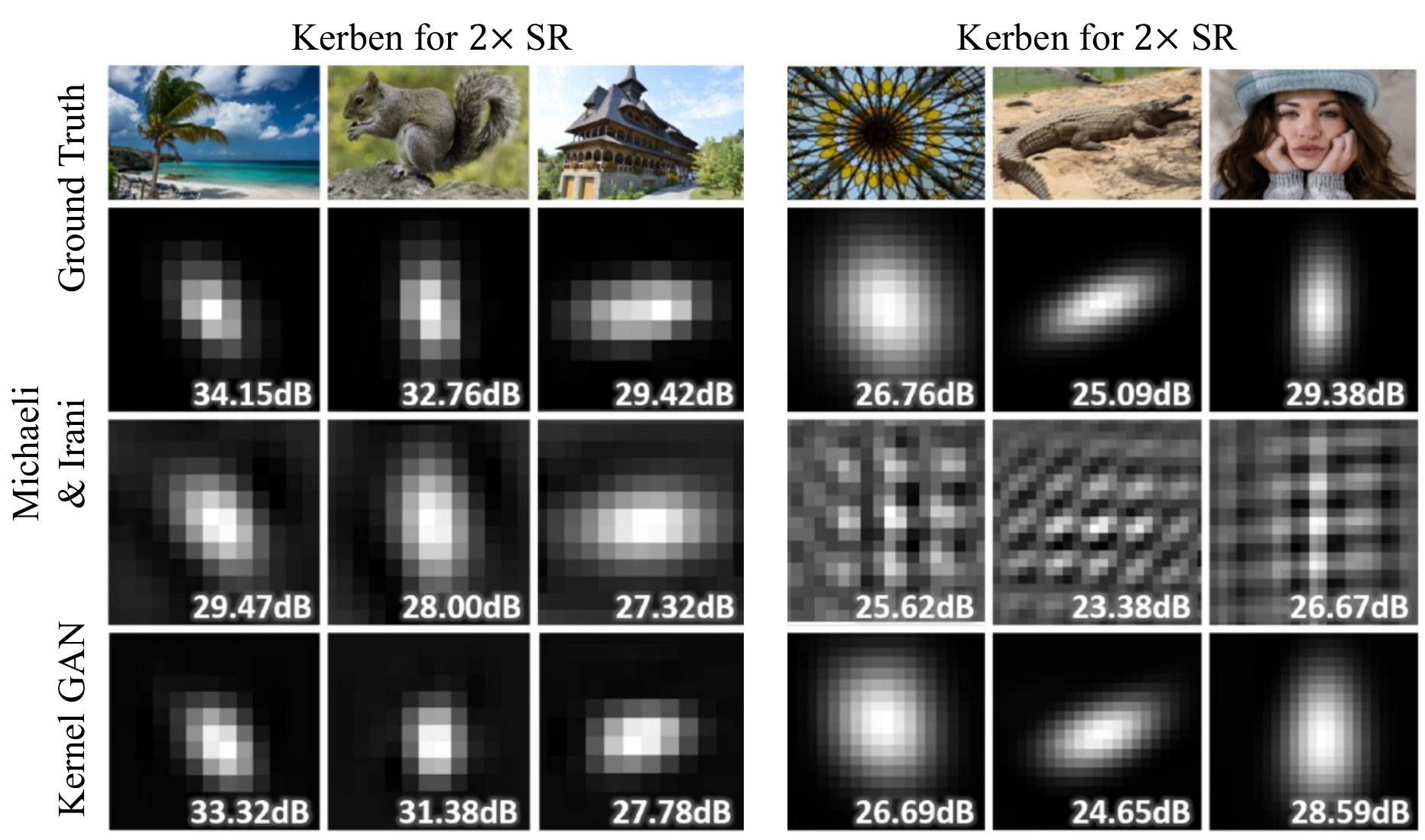}
    \caption{SR kernel estimation results of DNN method KernelGAN~\cite{bell2019blind} and non-DNN method~\cite{michaeli2013nonparametric}.}
    \label{fig:kernel}
    \vspace{-12pt}
\end{figure}

\begin{figure}[t!]
    \centering
    \includegraphics[width=0.4\textwidth]{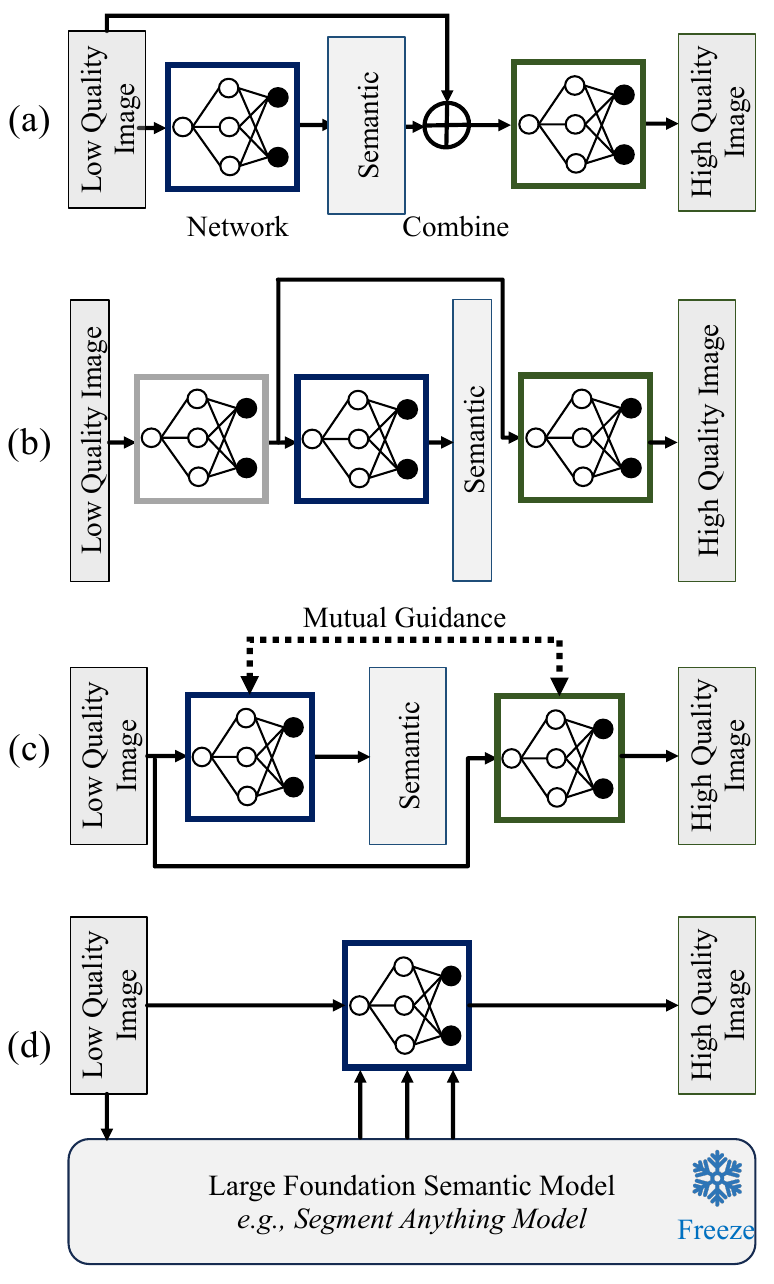}
    \caption{Different approaches to image restoration and enhancement with semantic information guidance. (a) the semantic output as the input for the image restoration and enhancement model. (b) the semantic segmentation model shares a backbone with the restoration and enhancement model. (c) enhancement and segmentation tasks are mutual guidance and alternatively optimized.
    (d) the large-scale foundation semantic segmentation model provides semantic information or features.
    (a)-(d) are simplified illustrations from \cite{wang2021towards}, \cite{wang2020dual}, \cite{li2022close}, \cite{jiang2023restore} respectively.}
    \label{fig:14-high-level-guide}
    \vspace{-6pt}
\end{figure}

\subsubsection{Explicit Modeling in DNN Structure Design}
\label{subsubsec:kernel_prior_1}
With such a degradation model, many works~\cite{zhang2018learning,xu2020unified,zhang2019deep,hui2021learning} focus on improving the performance of DNNs using the estimated kernel or noise information of the degradation models.
SRMD~\cite{zhang2018learning} is the first work that uses blur kernel and noise map as an additional input for image SR.
Specifically, SRMD proposes a stretching strategy using Principal Component Analysis (PCA) for projection on vectorized blur kernel to encode the kernel information.
Along this line, UDVD~\cite{xu2020unified} further proposes a dynamic convolutions block to extract features from the conditional input to generate per-pixel kernels for refinement.
Zhang \etal~\cite{zhang2019deep} introduced an unfolding optimization for image SR by considering more degradation information as the conditional input. They also turned the training target into a data sub-problem and a prior sub-problem, which can be alternately solved.
Similarly, FFDNet~\cite{zhang2018ffdnet} proposes to concatenate the noise-level map as the conditional input for image denoising.

In addition to the methods using pre-estimated degradation information, some works focus on estimating the kernel with DNN models and then combining the estimated kernel with the aforementioned deep restoration methods taking kernel information as conditional input.
KernelGAN~\cite{bell2019blind} proposes to use a deep linear network as the generator to parameterize the underlying kernel for image SR, while a discriminator is designed to distinguish the generated patches from the real LR image patches.
Once training is done, convolving all convolution filters in the generator can explicitly obtain the estimated kernel.
The learned kernel in KernelGAN can be more robust than non-DNN models as shown in Fig.~\ref{fig:kernel}.
FKP~\cite{liang2021flow} proposes a flow-based kernel modeling approach to generate reasonable kernel initialization and traverse the learned kernel manifold to achieve better kernel estimation results.

Furthermore, a series of works~\cite{gu2019blind,huang2020unfolding,guo2019toward} jointly estimate the degradation information and learn the target restoration tasks in an end-to-end manner.
ICK~\cite{gu2019blind} proposes an iterative kernel correction method to correct the estimated kernel information based on the SR results from the previous stage.
Huang \etal~\cite{huang2020unfolding} developed a deep alternating network by iteratively estimating the degradation and restoring an SR image.
CBDNet~\cite{guo2019toward} uses a noise estimation sub-network to predict the noise information and use it as the extra input to the DNN models for image denoising.

\vspace{-10pt}
\subsubsection{Explicit Modeling in Training Set Synthesis}
\label{subsubsec:kernel_prior_2}

Some other approaches also propose to cover more degradation patterns in the training dataset using more realistic kernels estimated from real images.
For instance, Zhou \etal ~\cite{zhou2019kernel} built a large kernel pool with data distribution learning based on some realistic SR kernels estimated from real-world LR images.
A similar strategy is employed in RealSR~\cite{ji2020real} to build a more generic training dataset using more realistic kernels.
Tran \etal~\cite{tran2021explore} proposed to use an encoder-decoder architecture to encode the blur kernels of the sharp-blurry image pairs into a blur kernel space and applied the learned kernel space to the blurry image simulation.
In the image denoising task, both GCBD~\cite{chen2018image} and C2N~\cite{jang2021c2n} use a GAN to generate realistic noise to build the paired dataset for image denoising.

\textbf{Discussion}:
In real-world scenarios, the degradation process of an image is usually unknown and can be affected by various factors beyond the pre-defined degradation model.
Therefore, explicitly modeling the kernel and noise information cannot yield satisfactory results for images whose degradation is not covered by these models.
Beyond the format of kernel and noise map, it is possible to use a more general way to encode the degradation information or supplementary information, \eg, learnable representation ~\cite{wang2021unsupervised} and data characteristic~\cite{kong2021classsr}.

\vspace{-10pt}
\section{Semantic Priors}

\label{primary-prior:semantic}
Semantic priors are based on the semantic information of the image, providing knowledge about the image content and high-level information.
We categorize semantic priors into two parts: (1) general semantic information as priors and (2) human-centric semantic information as priors.
General semantic information as priors (Sec \ref{sec:high_level_information_prior}) encompass various objects, scenes, and other visual elements.
Due to its significant research interest, the human-centric prior, \eg, body pose, facial prior,  represents a distinct form within the general semantic priors, warranting a separate presentation (Sec. \ref{sec:facial_prior}).

\subsection{General Semantic Information as Prior}
\label{sec:high_level_information_prior}

\noindent \textbf{Insight:} \emph{General semantic information refers to using semantic segmentation \cite{asgari2021deep} results or features as priors to guiding deep image restoration and enhancement.}

Semantic information can bring considerable constraints on the image contents, \eg, natural scene and the human body, face \cite{wang2021towards,liu2020connecting}, which could potentially provide adequate guidance for deep image restoration and enhancement.
Recently, many approaches have been proposed to use the semantic information as prior for the tasks, such as deblurring \cite{shen2018deep,shen2019human}, dehazing \cite{ren2018deep}, draining \cite{li2022close}, image/video SR \cite{wang2018recovering,wang2018high,rad2019srobb,wang2020dual}, low-light image enhancement \cite{fan2020integrating,zheng2022semantic,liang2022semantically} and combination of image deblurring and SR \cite{shen2018deep,shen2019human,chen2021progressive,purohit2021spatially}.
We roughly divide these methods into four categories, as shown in Fig.~\ref{fig:14-high-level-guide}.
The first three categories are divided according to the relationship between the image enhancement network and the semantic segmentation network, which can be trained with the image enhancement network simultaneously.
The last category of semantic information comes from a large foundation model, \eg, Segment Anything Model (SAM) \cite{kirillov2023segment}, which is a brand-new technology in deep learning and is currently booming.

A general illustration of the methods of the first category is shown in Fig.~\ref{fig:14-high-level-guide}(a), where the semantic information is taken as the input to DNNs for image restoration and enhancement.
Wang \etal~\cite{wang2021towards} presented the first low-light image enhancement framework utilizing the semantic information to estimate intrinsic reflectance parameters.
This can enhance the region details and guide the adaptive denoising based on the scene semantics.
Along this line, Ren \etal~\cite{ren2018deep} proposed a framework that takes the global semantic results as input to estimate the transmission maps in video dehazing.
Unlike these methods,
Shen \etal~\cite{shen2019human} and Chen \etal~\cite{chen2021progressive} directly took the semantic segmentation results as inputs to the image enhancement network. Specifically,
Shen \etal~\cite{shen2018deep} employed an encode-decode network with semantic information to accomplish the goal of image deblurring.
Chen \etal~\cite{chen2021progressive} replaced the encode-decode framework and proposed a semantic-aware style transformation framework for image restoration.
Although these methods have achieved good results by leveraging the semantic prior as guidance, they are sensitive to the errors accumulated from the semantic segmentation network.

\begin{table}[t!]
\caption{Application of semantic information as prior in image/video restoration and enhancement tasks}
\vspace{-13pt}
\label{tab:high_level_table}
\begin{center}
\begin{tabular}{ccccc}
\hline
Methods                             & Task                      & Highlight \\
\hline
\hline
TIP 2018         \cite{ren2018deep}            & \makecell[c]{Dehazing}                              & \makecell[c]{First work for semantic\\ in video dehazing}     \\
CVPR 2018        \cite{wang2018recovering}     & SR                                          & Textures recover \\
CVPR 2018        \cite{wang2018high}           & SR                                          & GAN-based \\
ICCV 2019        \cite{rad2019srobb}           & SR                                            & \makecell[c]{Object, background\\ and boundary} \\
CVPR 2020        \cite{wang2020dual}           & SR                            & Two-stream framework   \\
MM 2020      \cite{fan2020integrating}     & \makecell[c]{Low-light \&\\Denoise}               & \makecell[c]{First work for\\low-light enhancement} \\
TIP 2020         \cite{liu2020connecting}      & Denoise                                     & \makecell[c]{First work for\\ deep denoise} \\
MM 2022      \cite{li2022close}            & Deraining                                   & \makecell[c]{Bottom-up and \\top-down Paradigm} \\
\hline
\end{tabular}
\end{center}
\vspace{-15pt}
\end{table}

The methods of the second category jointly train the deep image restoration network and semantic segmentation network, which can relieve the impact caused by the accumulated errors, as shown in Fig.~\ref{fig:14-high-level-guide} (b).
Wang \etal~\cite{wang2020dual} proposed a representative framework by jointly learning a segmentation network and an SR network.
As they follow the same encoder-decoder structure and share the encoder, the SR image and SR segmentation results can be obtained from the LR input without any accumulated errors.
Similarly, Liu \etal~\cite{liu2020connecting} found that joint training of the image semantic segmentation network and denoising network can reinforce each other by improving the generalization capability in complex scenes.

The methods in the third category explore the collaborative relationship between the image restoration and enhancement tasks and high-level tasks, and introduce a mutual guidance strategy for the training of both networks, as shown in Fig.~\ref{fig:14-high-level-guide} (c).
For instance,
Li \etal~\cite{li2022close} proposed a framework containing a deraining network and a segmentation network.
The deraining network eliminates the image-level and feature-level rainy effects on the segmentation network, while the segmentation network learns the spatial-aware features to facilitate deraining by embedding the semantic information.
Similarly, Ma \etal~\cite{ma2020deep} proposed a joint human face SR and landmark detection framework based on mutual collaboration.
Furthermore, detection information can also be regarded as a special type of semantic prior, which can be applied to deep image restoration tasks~\cite{huang2020dsnet,shen2019human,ma2020deep}.
Shen \etal~\cite{shen2019human} exploited the detection information prior to distinguishing human and background, which guide the networks for pedestrian and background image deblurring, respectively.
The benefit is that prior can better drives networks for each task to focus on the specific regions containing either the pedestrian or background.

\begin{figure}[t!]
    \centering
    \includegraphics[width=0.48\textwidth]{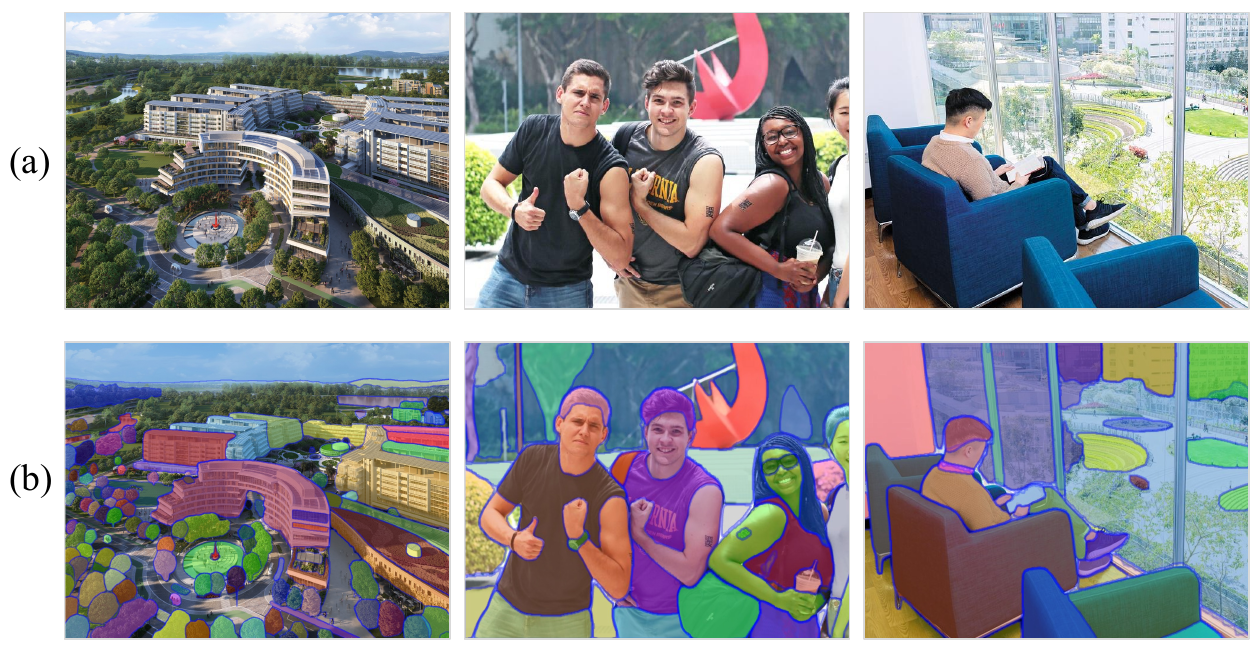}
    \caption{The segmentation results of SAM \cite{kirillov2023segment} on randomly selected images show good robustness and generalization.}
    \label{fig:sam-results-hkust}
    \vspace{-10pt}
\end{figure}

In recent times, large-scale foundation models \cite{radford2021learning,kirillov2023segment} have provided fresh impetus to the field of computer vision.
As an illustrative study, SAM \cite{kirillov2023segment} is trained in a promptable fashion on the largest segmentation dataset currently available.
The dataset consists of over one billion annotations across eleven million images.
The remarkable ability of SAM to segment anything provides novel inspiration for semantic-guided image enhancement, as shown in Fig.~\ref{fig:sam-results-hkust}.
Differing from the three categories as mentioned above, SAM-based methods \cite{jiang2023restore,xiao2023dive,zhao2023enlighten,lu2023can} acquire semantic information directly from the SAM pre-training model, in contrast to training the image enhancement network and segmentation network jointly.
For instance, Xiao \etal~\cite{xiao2023dive} presented a SAM prior tuning unit as an additional component within existing SR networks \cite{zhang2022accurate,chen2022cross}, aiming to incorporate semantic information from SAM as a guidance mechanism.
Nonetheless, these methods solely adopt a plug-in approach, overlooking the optimization of the image enhancement network backbone, thereby limiting the progress achieved.
As a powerful but emerging approach, other large-scale foundation models also hold substantial potential, further illustrated in the section on the large-scale foundation model (Sec. \ref{subsubsec:large-model}).

\textbf{Discussion:}
From the review, semantic prior can help improve the performance of deep image restoration \cite{wang2021towards}, while the restored outputs can help to learn the semantic segmentation networks \cite{liu2020connecting}.
Therefore, avoiding the dependencies between tasks and exploiting more efficient mutual collaboration strategies deserve more research.
On the other hand, the utilization of large-scale foundation models \cite{kirillov2023segment} holds significant potential, as it can furnish steadfast and resilient semantic information for guidance.
Moreover, based on the theoretical analysis in Sec.~\ref{sec:definition}, semantic prior boosts the deep image enhancement model capacity $f_\Phi$ by supplying extra semantic information, and they can be easily extended to other fields, \eg, HDR imaging, and image compression artifact removal.
Lastly, it is also promising to explore fusing the high-level tasks and low-level tasks via the semantic prior.

\begin{table*}[t!]
\centering
\caption{Application of facial prior and human body prior in the deep image restoration tasks.}
\resizebox{\linewidth}{!}{
\begin{tabular}{ccccc}
\hline
Method& Publication& Task & Prior& Highlight\\
\hline
\hline
Li~\cite{li2018learning}&ECCV 2018&  Face Restoration& Reference-based Prior & High-quality guided image of the same identity\\
Yu~\cite{yu2018face} &ECCV 2018&  Face Hallucination (SR)& Facial Component & Facial component heatmaps\\
Chen~\cite{chen2018fsrnet}&CVPR 2018 & Face Hallucination (SR) & Geometry & Facial Landmarks / Parsing Maps\\
Bulat~\cite{bulat2018super}&CVPR 2018 & Face Hallucination (SR) & Geometry & End-to-end face SR and landmark localization\\
Chrysos~\cite{chrysos2019motion}&IJCV 2019&  Face Deblurring&Geometry& Facial Landmarks / Parsing Maps\\
Kim~\cite{kim2019progressive}&BMVC 2019 &  Face Hallucination (SR)& Geometry& Facial Landmarks\\
Li~\cite{li2020blind} &ECCV 2020&  Face Restoration& Reference-based Prior & Deep face components dictionaries\\
Ma~\cite{ma2020deep}&CVPR 2020&  Face Hallucination (SR)& Geometry& Facial Landmarks / Parsing Maps\\
Li~\cite{li2020enhanced}&CVPR 2020&  Face Hallucination (SR)& Reference-based Prior & High quality images as reference \\
Kalarot~\cite{kalarot2020component} &WACV 2020&  Face Hallucination (SR)& Facial Component & Facial component-wise attention maps\\
Shen~\cite{shen2020exploiting}& IJCV 2020& Face Deblurring& Facial Component & Semantic labels priors and local structures prior \\
Yasarla~\cite{yasarla2020deblurring}&TIP 2020 &  Face Deblurring& Facial Component & Semantic labels priors\\
Grm~\cite{grm2019face} & TIP 2020 &  Face Hallucination (SR)& Recognition Model& Identity prior from face recognition model \\
Kim~\cite{9506610}&ICIP 2021&  Face Hallucination (SR)& Facial Component & Non-parametric facial prior\\
Chen~\cite{chen2021progressive} &NeurIPS 2021&  Face Restoration& Parsing Maps& Semantic aware style transformation\\
Hu~\cite{9591403}&TPAMI 2021 &  Face Restoration& Geometry& Plug-and-play 3D facial prior \\
Chen~\cite{chen2021progressive} &CVPR 2021&  Face Restoration& Geometry& Parsing Maps\\
Wang~\cite{wang2021towards} &CVPR 2021&  Face Restoration& GAN Facial Prior & Prior in a pre-trained face GAN\\
Yang~\cite{yang2021gan} & CVPR 2021&  Face Restoration& GAN Facial Prior & Fine-tune the GAN prior embedded DNN\\
Jung~\cite{jung2022deep} & WACV 2022&  Face Deblurring& Recognition Model& Deep features of face recognition networks \\
\hdashline
Shen~\cite{shen2019human} & CVPR 2019 & Human Motion Deblur & Human body detection & Disentangles humans and background \\
Lumentut~\cite{lumentut2020human} & ACCV 2020 & Human Motion Deblur & Localized body prior & First adversarial framework from human prior \\
Liu~\cite{liu2021accurate} & TIP 2021 & Human-centric Image SR & Human body parsing & Human body prior estimation branch \\
\hline
\label{tab:face}
\end{tabular}}
\vspace{-15pt}
\end{table*}

\begin{figure}[t]
    \centering
    \includegraphics[width=0.89\linewidth]{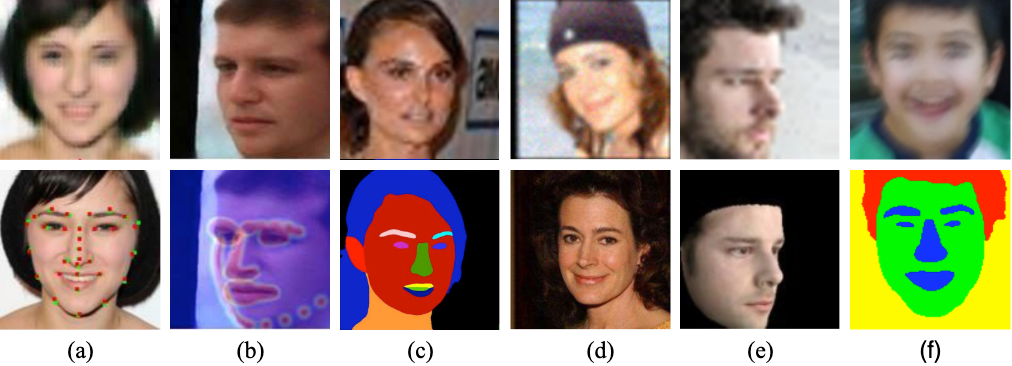}
    \caption{Visual examples for some facial priors. (a) Facial landmarks prior~\cite{chen2018fsrnet}; (b) Facial component prior~\cite{yu2018face}; (c) Facial parsing map prior~\cite{chen2021progressive}; (d) Reference-based (high-quality image) prior~\cite{li2018learning}; (e) 3D facial prior~\cite{9591403}; (f) Semantic label prior~\cite{yasarla2020deblurring}.}
    \label{42facial}
    \vspace{-8pt}
\end{figure}

\subsection{Human-centric Prior}
\label{sec:facial_prior}

\noindent \textbf{Insight:} \textit{
Features such as the unique geometry and structure of the human body or face, as shown in Fig.~\ref{42facial}, can be utilized to guide deep image restoration and enhancement.
Facial priors received more attention, and human body-based priors have also been researched.}

In this section, we present a comprehensive overview of facial priors, which have received substantial attention, followed by an analysis of priors about the human body. Despite the relatively limited number of studies investigating human body priors, the burgeoning interest in virtual reality (VR) has sparked a focus on these priors for enhancing images containing humans to augment performance in downstream tasks \cite{zheng2019deephuman,jin2021pedestrian}.

Facial prior can be divided into three types~\cite{wang2021towards}: face-specific prior~\cite{yu2018face,chen2018fsrnet,yasarla2020deblurring}, reference-based prior~\cite{li2018learning,li2020blind,li2020enhanced}, and generative facial prior~\cite{wang2021towards,yang2021gan}.
The widely used face-specific prior is based on the common sense that human faces are in a controlled setting with small variations. Consequently, the unique geometry and spatial distribution information, \eg, facial landmarks~\cite{ma2020deep}, can be utilized as geometric facial prior~\cite{chen2018fsrnet,bulat2018super,kim2019progressive,9591403,chen2021progressive} to accomplish the face image restoration tasks. Meanwhile, various features of each component, \eg, the eyes, can be used as prior knowledge.
Typically, FSRNet~\cite{chen2018fsrnet} is a representative framework consisting of two stages. In the first stage, a prior estimation network is designed to provide the facial landmark heatmaps and parsing maps simultaneously. In the second stage, a restoration network, which takes the heatmaps and parsing maps as inputs, is proposed to restore a clean face image.
However, applications of face-specific prior are still limited because the low-quality images often lose textural information about the facial details.

The reference-based priors rely on guidance from high-quality face images~\cite{li2020enhanced,li2018learning} or facial component dictionaries~\cite{li2020blind}. These priors are always obtained by applying traditional methods, \eg, K-means, to cluster the perceptually significant characteristics of facial components. These priors are then applied to tackle more challenging face image restoration problems.

The generative facial prior is the most extensively employed method, capable of furnishing copious reference information.~\cite{chen2021progressive,yang2021gan,wang2021towards}.
For instance, GFP-GAN~\cite{wang2021towards} leverages the facial distribution captured by a pre-trained GAN as a facial prior to achieving joint restoration and color enhancement. Meanwhile, some specific facial priors are proposed to constrain the DNN training procedure. For instance, the identity prior~\cite{grm2019face} leverages the pre-trained recognition network models to provide the auxiliary identification information for face hallucination. Lastly, whether these specific facial priors can be further extended to more image enhancement and restoration tasks is still worth exploring.

The research focused on utilizing human body information also attracts attention. Notably, three representative works have emerged. Liu et al. \cite{liu2021accurate} introduced parsing information from different parts of the human body (\eg, arm, shirt) into the SR network backbone; Shen \etal~\cite{shen2019human} utilized human body detection information to distinguish humans from the background, designing a network with multi-branches that separately processes foreground and background to address motion blur related to human movement; Lumentut \cite{lumentut2020human} employed human body location information to guide network attention specifically towards human body blurring.
Human-centric image enhancement techniques leverage the semantic disparity between human body regions and background, allowing networks to enhance performance by discerning distinct regions.

\textbf{Discussion:}
Several studies exploit human-centric priors in SR and deblur, effectively integrating prior knowledge of the human body and face to endow the network with valuable geometric and semantic information.
SR and deblur techniques assume a pivotal role in diverse downstream tasks within human-centric scenes, encompassing pose estimation, person re-identification, and VR.
Given the current widespread adoption of VR applications, particular attention should be directed toward tasks involving human-centric image enhancement. In future endeavors, valuable insights from the face prior applications can be effectively leveraged to incorporate 3D and other priors, guiding the development of human-centric image enhancement techniques.

\vspace{-10pt}
\section{Deep Learning-based Prior}
\label{primary-prior:deep-learning}

Deep learning-based prior are learned from deep learning models, \eg, CNN, using the structure and parameters of deep neural networks to restore and enhance images.
We introduce this prior in two distinct parts: the deep image prior (Sec. \ref{sec:deep_image_prior}) and the pre-trained model as a prior (Sec. \ref{subsec:pretrain}).

\begin{table}[t!]
\centering
\caption{Representative works with deep image prior.}
\setlength{\tabcolsep}{1mm}{
\resizebox{0.8\linewidth}{!}{
\begin{tabular}{ccc}
\hline
 Method & Task & Highlight \\
\hline
\hline
Ulyan~\cite{Ulyanov_2020_dip} & SR;Denoising  & First Work \\
Yosef~\cite{d-dip} & Dehazing  &  Coupling multiple DIP \\
Ren~\cite{ren2020neural} & Deblurring  & \makecell[c]{Generating blur \\kernel using DIP}\\
Heckel~\cite{heckel2018deep} & Denoising  & \makecell[c]{Lightweight network\\ for DIP} \\
Cheng~\cite{cheng2019bayesian} & Denoising  & \makecell[c]{Bayesian optimization\\ approach for DIP} \\
Jo~\cite{Jo_2021_ICCV} & Denoising  & \makecell[c]{Stop criteria of\\ DIP training} \\
Arican~\cite{ISNAS-DIP} &SR;Denoising & \makecell[c]{Neural architecture\\ search for DIP}\\
Lei~\cite{dvp_lei2020blind} & Video Dehazing & DIP on video\\
\hline
\label{tab:dip}
\end{tabular}}}
\end{table}

\vspace{-10pt}
\subsection{Deep Image Prior}
\label{sec:deep_image_prior}

\noindent \textbf{Insight:} \textit{A randomly initialized CNN can generate high-quality images based only on the degraded images. The structure of a CNN can serve as the prior. }

\begin{figure}[t!]
\centering
\includegraphics[width=0.5\textwidth]{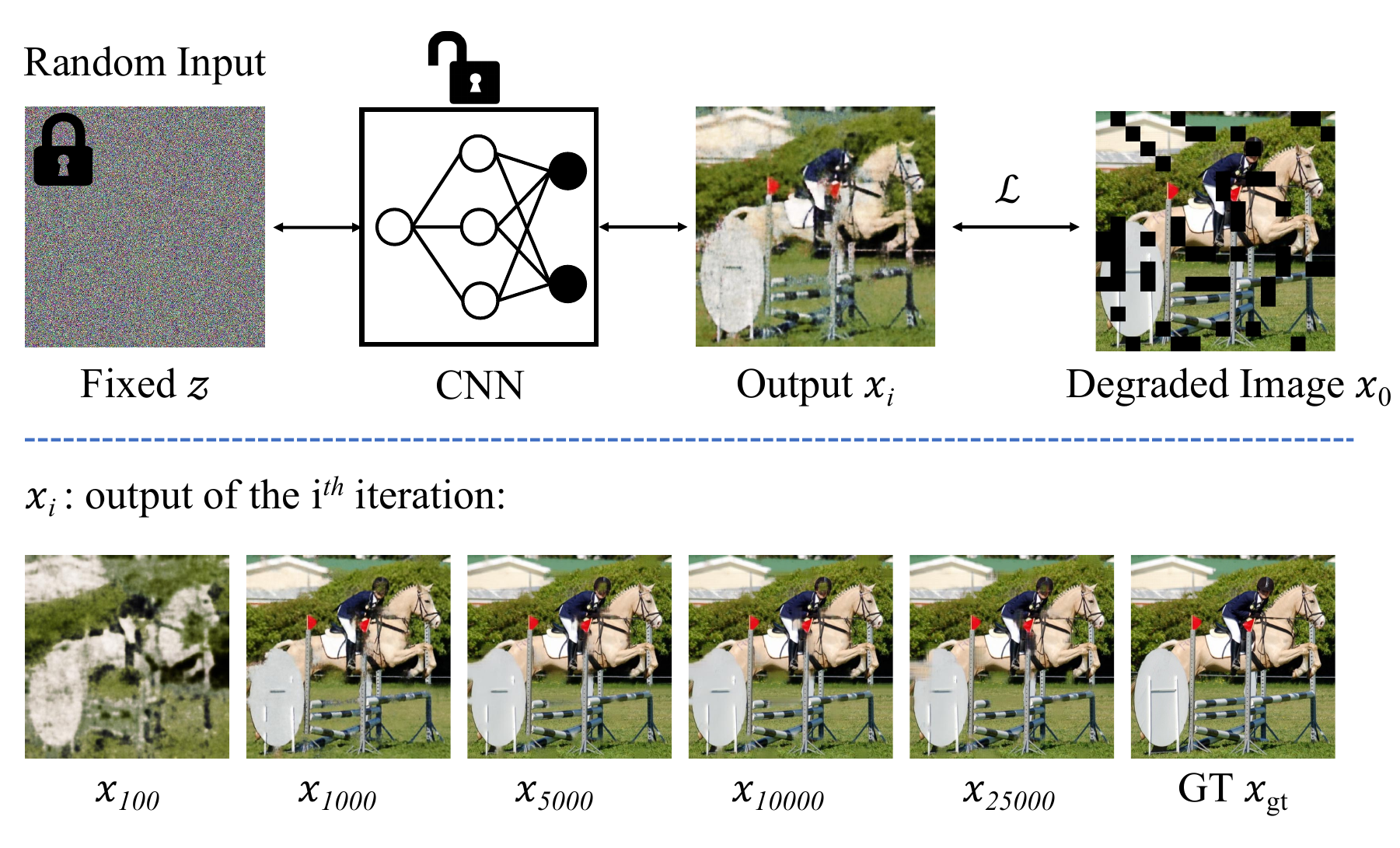}
\caption{Pipeline and outputs of DIP in image inpainting task.
The CNN is only trained on the random and fixed input $z$, and typically, $\mathcal{L}$ is vanilla MSE loss between $x$ and $x_0$. We reproduce DIP with U-Net and show how output $x$ evolves through iterations. Some details of $x$, e.g., the sign, deteriorate with larger iterations (overfitting). Some early stopping principles are required to obtain the optimal restored output $x^*$, or it needs to be manually selected.}
\label{17-dip}
\centering
\vspace{-8pt}
\end{figure}

CNNs have shown strong capability in image processing tasks. Ulyanov \etal~\cite{Ulyanov_2020_dip} found that the structure of a CNN can inherently capture the natural image distribution.
Specifically, it is demonstrated that a network needs no external datasets to train on and can merely rely on itself to restore or enhance images. Therefore, CNNs itself is a prior, dubbed Deep Image Prior (DIP).
Given a randomly initialized CNN parameterized by $\theta$, DIP runs an optimization process that recovers the observed image from a noisy input through $\theta$ by minimizing a task-dependent loss function $\mathcal{L}$, as shown in Fig.~\ref{17-dip}. In most tasks, the $L_2$ distance energy function is typically used to compare the generated image $x$ with degraded one $x_0$.
Gradient descent-based methods are applied to optimize $\theta$ to produce a favorable restored image $x^*$.

DIP is a universal prior probably due to the inherent image-level inductive bias comes with the CNN. that can be easily applied to nearly all image processing tasks, such as SR~\cite{Ulyanov_2020_dip}, deblurring~\cite{ren2020neural}, dehazing~\cite{d-dip}, denoising~\cite{dip_noise2void_2019_CVPR}.
The key of DIP is its compatibility to many tasks and impressive performance.
DIP is highly generalizable because it does not rely on handcrafted kernels or equations, and hence significantly mitigates the notorious hyperparameter-tuning problem.
The most popular ways to exploit DIP probably is designing novel network structure and preventing network from catastrophic overfitting. These two important lines of research will be introduced below.

\subsubsection{Network Structures for Deep Image Prior}
The variety and flexibility of neural network enable rich potential of DIP.
Heckel \etal~\cite{heckel2018deep} proposed to ease the heavy parameters of CNNs in DIP and discovered that a lightweight non-convolutional neural network is sufficient to generate high-quality images.
SelfDeblur~\cite{ren2020neural} argues that DIP is limited to capturing the prior of blur kernels as it is designed to generate natural images. Thus, it adopts an additional fully-connected network to model the blur kernel.
Different tasks or even images often need different network structures to achieve the best performance. Therefore, some research try to automate the search of suitable architecture. \cite{ISNAS-DIP} introduced neural architecture search into DIP and added novel components to the search space, such as upsampling cells and cross-scale residual connections. Despite the large search space size, they significantly reduce search time by adding image-specific metrics.

DIP restores the whole image end-to-end, acting as a black box. To visualize and make it more explicit, another line of research combines multiple CNNs together so that each image can be decomposed into several layers according to tasks. Yosef \etal~\cite{d-dip} first proposed Double-DIP (D-DIP) that decomposes images into several basic components. D-DIP uses one DIP generator network to construct one layer based on the fact that DIP can capture the entire image, which is more complex than one layer. These decouple-based methods can be easily applied to downstream tasks. For example, in the dehazing task, Li \etal~\cite{zs_image_dehazing} decomposed a hazy image into a hazy-free image layer, transmission map layer, and atmospheric light layer. Most recently, He \etal~\cite{He_2023_dap} proposed to add self-attention to the network and named it Deep Attention Prior. The emerging pixel correspondence facilitates better interpretability.

\subsubsection{Preventing Overfitting in Deep Image Prior}
DIP has a basic principle: optimizing a network $f_\theta$ to produce the desired output $x^*$.  Notwithstanding, the optimization is solely based on the observed image $x_0$, and such single-image training poses a huge risk of overfitting. In whatever tasks, to generate better images, it is essential to prevent overfitting or take a timely interruption during the training. Otherwise, the output may converge to the degraded target $x_0$ or collapse.

Cheng \etal~\cite{cheng2019bayesian} theoretically studied DIP in a Baye-sian approach. They showed that replacing stochastic gradient descent with stochastic gradient Langevin dynamics can significantly reduce overfitting and yield better results in denoising and inpainting tasks.
Quan \etal~\cite{dip_dropout_2020_CVPR} proposed to use pixel-level dropout based on Bernoulli sampling to reduce overfitting and prediction bias when denoising.
Besides, Jo \etal~\cite{Jo_2021_ICCV} monitored the DIP denoising process by the effective degrees of freedom to impose an in-time stop.

Priors can guide the search to high-quality images, and DIP is flexible to be combined with other priors.
Here, other image priors serve as explicit regularizations for DIP. For one thing, some downstream tasks will be benefited from the given priors; for another, these priors can robustify DIP and mitigate overfitting.
This kind of method usually integrates an existing prior as a term in the loss function during training. For instance, \cite{mataev2019deepred} applied the regularization by denoising, and Liu \etal~\cite{liu2019DIP_TV} applied the total variation prior. Alternatively, Li \etal~\cite{Li_2023_CVPR} proposed to optimize the initial noise with the network frozen; MetaDIP \cite{zhang2022metadip}
showed that initializing the network with pre-trained weights also make the optimization shorter and more robust.

\vspace{-5pt}
\subsubsection{Deep Video Prior}
Videos consisting of a sequence of images can also benefit from DIP. Lei \etal~\cite{dvp_lei2020blind} propose Deep Video Prior (DVP) and show that the distance between correspondences in the output frames can be minimized internally by a CNN. It is based on the fact that given a CNN, similar patches yield similar results at the early stage of training. Therefore, DVP achieves convincing temporal consistency between output frames, and no explicit regularization term needs to be hand-crafted.

\textbf{Discussion:}
DIP hits the literature by its simpilicity and generalizability. It provides new insight into low-level vision tasks and ignites research into exploring the property of deep models. DIP is an effective unsupervised approach that is universally applicable, usually not constraint by the image size, content, or quality. Despite that vanilla DIP still falls short when compared with many supervised counterparts, its task-specific derivatives are close to state-of-the-art supervised methods. The optimization process of DIP is image-specific. Though DIP is free from large-scale training, it usually takes thousands of iterations to produce one satisfying image, consuming considerable time and computation resources. In addition, DIP always tends to overfit the degraded images. Researchers are still actively seeking solutions for the two remaining problems.

\vspace{-8pt}
\subsection{Pre-trained Model as Prior}
\label{subsec:pretrain}
\noindent \textbf{Insight:} \textit{The pre-trained models often contain specific knowledge of the high-quality images or low-quality images, which can be used as generators or regularization terms for deep image restoration and enhancement.
}

\begin{figure}
    \centering
    \includegraphics[width=0.45\textwidth]{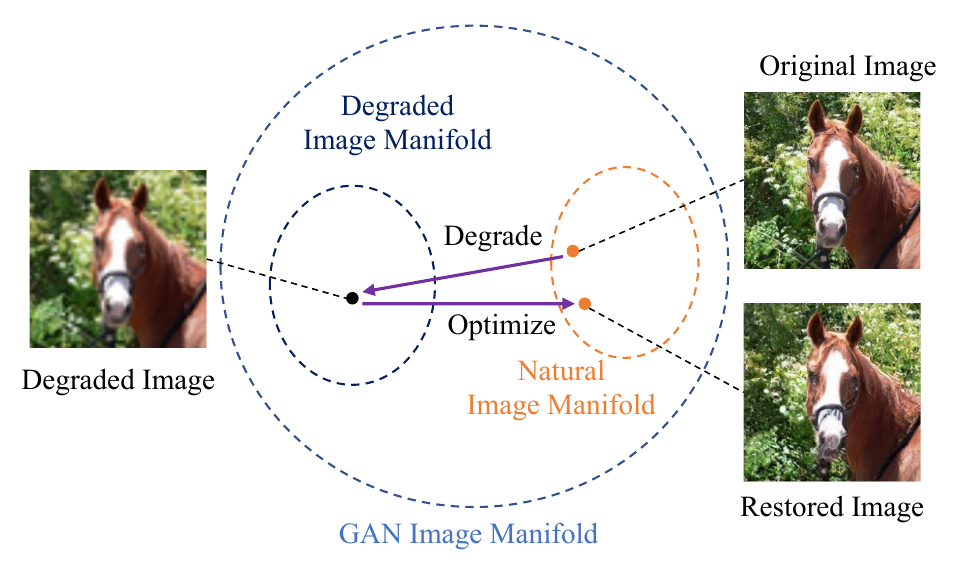}
    \caption{Illustration of using GAN inversion as prior. In the image manifold of the pre-trained GAN, we can search the latent code for the degraded image that matches the corresponding original natural image and generate the restored image with the code.}
    \label{fig:22-gan-inversion}
    \vspace{-12pt}
\end{figure}

\subsubsection{GAN Inversion as Prior}
\label{subsubsec:pretrain1}
GAN has been dominant in image generation~\cite{richardson2021encoding,alaluf2021only_ganprior} and editing tasks~\cite{richardson2021encoding,tewari2020stylerig}. A well-trained generator can theoretically reproduce any image, given a corresponding latent code ${z}$. Therefore, pre-trained GAN models, such as StyleGAN~\cite{style_gan}, can be used as a prior, i.e., GAN prior.
GAN prior is based on GAN-inversion. It first maps the degraded images to unique latent codes and then searches for the preferable codes in the latent space.
Finally, restored or enhanced images can be directly produced out of the optimized codes, as shown in Fig.~\ref{fig:22-gan-inversion}.
In short, For any degraded image $\hat{x}$ and its latent code $\hat{z}$, the nature image ${x}$ can be obtained by finding a natural latent code ${z}$ in the output space of the pre-trained generator $G$.

PULSE \cite{PULSE_2020_CVPR} firstly introduces a GAN prior for face SR by proposing an efficient searching strategy in the HR facial image manifold.
As the latent code is a feature vector with limited dimensions, it might fail to capture rich spatial information. To this end, multi-code GAN prior~\cite{mGAN_2020_CVPR} proposes to use multiple latent codes simultaneously to process one image.
They adaptively fuse the features generated by latent codes at a certain intermediate layer of $G$. With such a more sophisticated model, GAN prior is able to have broader applications, such as colorization, denoising, and inpainting.

However, Pan \etal,~\cite{deep_generative_prior_dgp} discovered that gaps exist between the image manifold captured by the generator $G$ and an input image from the wild, as $G$ is likely to overfit its training dataset. To address this issue, they proposed deep generative prior (DGP) that simultaneously updates $G$ and latent code ${z}$, showing noticeable improvement in various restoration tasks.
Another strategy to better leverage GAN prior is training an encoder $E(x;\theta_E)$ beforehand to produce image-specific, in-distribution $z$ to feed to $G$. GLEAN~\cite{GLEAN_GANprior_2021_CVPR} appends a trainable encoder before $G$ and a decoder after $G$ for the SR task. Similarly, in face restoration, GFP-GAN \cite{wang2021towards} adds a degradation removal module and obtains high-quality multi-level spatial features dedicated to $G$. These kinds of pre- or post-processing can further exploit the potential of GAN prior.

\vspace{-10pt}
\subsubsection{Generative Prior for Training Set Synthesis}
\label{subsubsec:pretrain2}
GAN can also be used to exploit data distribution learning and serve as a degradation prior to generating the paired synthetic training set.
Specifically, the GAN-based method pre-trains a generator model for the degradation process by using one or more discriminators to distinguish generated image samples from real ones.
Following this fashion, Degradation GAN~\cite{bulat2018learn} is a representative framework that combines a pre-trained degradation generator to generate LR images from HR images and proposes an SR network to super-resolve the LR images.
To reduce the domain gap between the generated LR images and real-world LR images, DASR~\cite{wei2021unsupervised} proposes to use both types of images to train the SR model.
Zhang \etal,~\cite{zhang2020deblurring} proposed the first framework that learns how to blur sharp images with an unpaired sharp and blurry image dataset and then generates the image pairs for training the deblur model.


\subsubsection{Deep Denoiser Prior for Model-based Methods}
\label{subsubsec:pretrain3}
Owing to the interpretability and adaptivity of classic model-based methods, several image restoration methods propose to integrate DNNs into the Maximum A Posteriori (MAP) framework with iterative optimization algorithms as a regularization prior.
Given a clean image $x$ and a degraded image $y$, the MAP framework minimizes an energy function composed of a data term and a regularization term.
The data term keeps the solution consistent with the degradation process, while the prior term enforces desired property on the solution to alleviate the ill-posed problem.
Based on this framework, one representative work proposed by Zhang \etal,~\cite{zhang2017learning} integrates a set of fast and effective CNN denoisers into the prior term to solve various restoration tasks, such as denoising, deblurring, and super-resolution.
The extended work~\cite{zhang2021plug} proposes to use a stronger U-Net denoiser to achieve better performance and apply it to more restoration tasks.
A similar idea is also adopted in \cite{dong2018denoising} by replacing the prior term with deep denoiser models.
Moreover, deep unfolding network~\cite{zhang2020deep1} proposes to optimize all parameters, including the denoiser prior term in an end-to-end manner, delivering better performance.

\begin{table}[t!]
\caption{Application of large-scale foundation models as prior in image/video restoration and enhancement tasks.}
\vspace{-15pt}
\label{tab:high_level_table}
\begin{center}
\setlength{\tabcolsep}{1mm}{
\resizebox{\linewidth}{!}{
    \begin{tabular}{ccc}
\hline
    Method & Task & Highlight \\
\hline
\hline
    \cite{abu2022adir} & SR;Deblurring & CLIP\\
    \cite{tan2023exploring} & Deraining;Dehazing;Desnowing & CLIP\\
    \cite{bai2023textir} & SR & CLIP\\
    \cite{wang2023exploiting} & SR & Stable Diffusion\\
    \cite{jiang2023restore} &  Deblurring;Denoising & SAM\\
    \cite{xiao2023dive} & SR;Denoising & SAM\\
    \cite{zhao2023enlighten} & Low-light & SAM\\
    \cite{lu2023can} & VSR & SAM\\
    \cite{jin2023let} & Dehazing & SAM\\
\hline
\end{tabular}}}
\vspace{-8pt}
\end{center}

\end{table}

\begin{figure}[t!]
\centering
\includegraphics[width=0.45\textwidth]{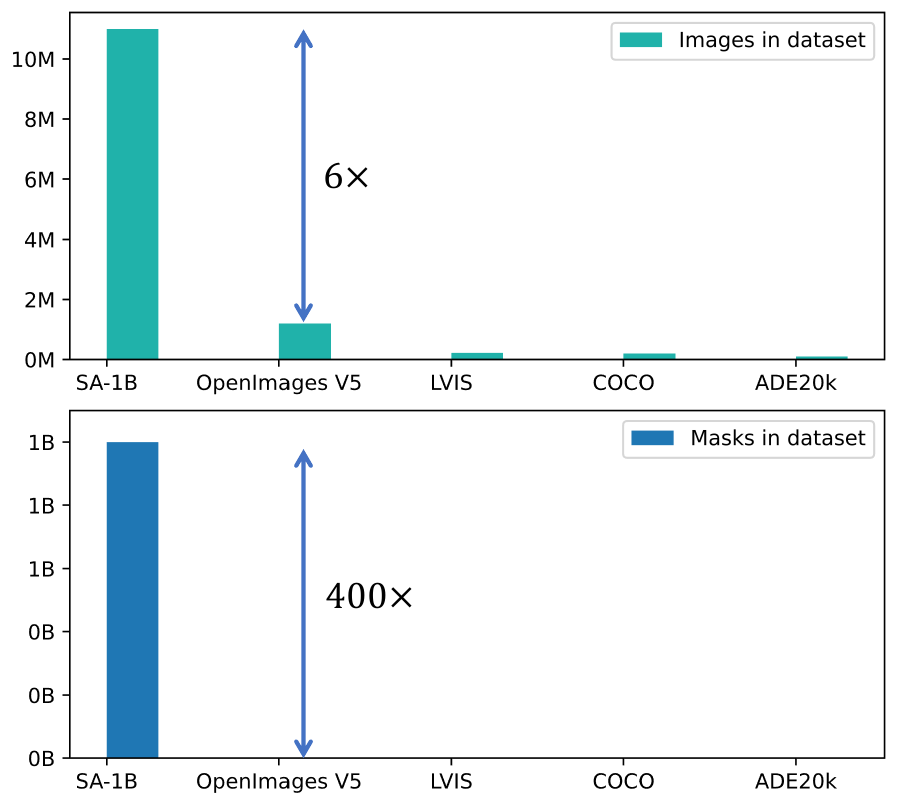}
\caption{The count of image and mask in the SAM dataset \cite{kirillov2023segment} and previous datasets OpenImage-v5 \cite{kuznetsova2020open}, LVIS \cite{gupta2019lvis}, COCO \cite{lin2014microsoft}, ADE20k \cite{zhou2019semantic}. The numbers of masks in other datasets are too small to be shown.}
\label{fig:1-SAM-1B}
\centering
\vspace{-10pt}
\end{figure}

\subsubsection{Large-scale Foundation Model as Prior}
\label{subsubsec:large-model}
Large-scale foundation models~\cite{bommasani2021opportunities}, or large-scale pre-trained models, offer substantial prior knowledge for diverse image restoration and enhancement tasks by exploiting the learned rich representations from extensive training data.
Due to their robust discriminative or generative abilities, large-scale foundation models, such as connecting text and images (CLIP)~\cite{radford2021learning}, diffusion models~\cite{rombach2022high}, and the segment anything model (SAM)~\cite{kirillov2023segment}, have gained traction in image enhancement tasks, such as SR, denoising, and deblurring.
However, adapting these models to image enhancement tasks poses challenges due to the demanding requirements for high-fidelity results.
To address this, researchers develop novel approaches \cite{abu2022adir,jin2023let,zhao2023enlighten} that utilize the strengths of foundation models while customizing them for specific tasks.


CLIP~\cite{radford2021learning} is a large-scale foundational model that effectively acquires visual concepts through natural language supervision, utilizing a dataset comprising over 400 million image-text pairs.
It can recognize various objects and scenes across different datasets without explicit training.
Therefore, ADIR~\cite{abu2022adir} uses CLIP as a versatile retrieval model to retrieve semantically similar images as references for image restoration tasks.
TextIR~\cite{bai2023textir} framework harnesses the aligned text-image feature of CLIP, enabling flexible and controllable restoration through text descriptions. Simultaneously, \cite{tan2023exploring} designs a model for adverse weather image restoration, incorporating a spatially-adaptive residual encoder and a CLIP weather prior embedding module to deal with varying weather conditions based on the text of weather.

Stable Diffusion \cite{rombach2022high} is a latent text-to-image diffusion model that has been trained on a vast dataset comprising 2.3 billion English-captioned images.
This model exhibits the capability to generate photo-realistic images when provided with textual input.
Therefore, it garners significant attention within the image restoration domain.
A blind SR approach~\cite{wang2023exploiting} exploits these models, featuring a time-aware encoder, controllable feature wrapping module, and progressive aggregation sampling strategy to overcome size constraints, and has demonstrated excellent performance on synthetic and real-world benchmarks.

Lastly, for the segmentation foundation model, SAM \cite{kirillov2023segment} is used as a prior in image restoration tasks by incorporating segmentation information, particularly when degradation relates to the semantic content of an image~\cite{jiang2023restore,xiao2023dive,zhao2023enlighten,lu2023can}.
The training dataset of SAM currently holds the distinction of being the largest segmentation dataset, encompassing over 10 million images and 1 billion labels, as shown in Fig. \ref{fig:1-SAM-1B}.
The vast scale of this dataset surpasses that of previous datasets, thereby enabling the SAM model to exhibit remarkable generalization performance and robustness, which can bring great opportunities to image enhancement fields.
For instance, an interactive image restoration approach~\cite{jiang2023restore} has been developed that allows users to choose from a range of results and operates at a per-object level by mask generated from SAM to generate different outcomes.
In \cite{xiao2023dive}, SAM provides the prior knowledge,\eg semantic masks, as extra features to boost the performance of the existing image restoration model.
Similarly, Enlighten Anything~\cite{zhao2023enlighten} fuses the semantic masks from SAM with low-light images to enhance visual perception.
In video SR, the semantic masks can be used as guidance information for frame alignment and fusion.
Moreover, SAM shows great potential~\cite{jin2023let} in detecting gray-scale segmentation masks from foggy images, which can provide prior knowledge for the existing dehazing model.
This suggests that combining segmentation and restoration in a unified pipeline can greatly improve the quality of restored images.
Nevertheless, due to the novelty of SAM, its full potential value remains unexplored.
Discussions of further direction regarding SAM are presented in Sec. \ref{subsubsec:potential-foundation}.
We hope this article can ignite novel ideas for the foundation model for image enhancement.

\textbf{Discussion:}
The rationale behind employing a pre-trained model as prior lies in the effectiveness of the knowledge acquired from a large-scale dataset, which serves as a valuable constraint and guide for image enhancement.
In simple terms, the efficacy of the constraints and guidance provided by the pre-training data set increases with its larger size.
Recently, the advent of large language and visual foundation models has instigated a new wave of revolution within the deep learning domain.
Utilizing these large-scale pre-models presents a valuable research avenue within the realm of image enhancement, meriting active exploration by researchers.

\begin{table*}[t!]
\centering
\caption{Application of prior in specific deep image restoration and enhancement tasks. We mark some promising areas that have not been studied enough as future directions. The definitions of $f_\Phi$, $p_l$, $p_h$ are in Sec. \ref{sec:definition}. $\uparrow$ means that prior can increase $f_\Phi$, $p_l$, $p_h$. $\star$ means suggested potential direction, which will be discussed in Sec. \ref{sec:future_direction}.}
\label{tab:summarize_priors}
\resizebox{\linewidth}{!}{
\begin{tabular}{c|ccccc}
\hline
Prior\textbackslash{}Task & 
Super Resolution  & 
Deblurring            & 
\makecell[c]{Dehazing;\\ Deraining;\\  Desnowing} & 
Denoising & 
\makecell[c]{HDR;\\Low-light \\enhancement}
\\
\hline
\hline

\makecell[c]{Physics-based\\ Prior} &
\textcolor{blue}{\textit{Future Direction}}&
\makecell[c]{ASM \\ $f_\Phi\uparrow$} &
\makecell[c]{ASM; \\ Rain Model \\ $f_\Phi\uparrow$}&
\textcolor{blue}{\textit{Future Direction}}&
\makecell[c]{Retinex Model \\ $f_\Phi\uparrow$}
\\

\hline

Temporal Prior &
\makecell[c]{Optical Flow(video) \\ $f_\Phi\uparrow$} &
\makecell[c]{Temporal Sharpness; \\ Optical Flow(video) \\ $f_\Phi\uparrow$} &
\makecell[c]{Optical Flow \\ (Video) \\ $f_\Phi\uparrow$}& 
\makecell[c]{Optical Flow \\ (Video)}&
\makecell[c]{Optical Flow \\ (Multi-Exposure) \\ $f_\Phi\uparrow$} 
\\
\hline
\makecell[c]{Statistical Image \\ Feature as Prior}     & 
\makecell[c]{Gradient Guidance \\ $p_h\uparrow, f_\Phi\uparrow$} & 
\makecell[c]{Extreme Channel Prior; \\ Two-tone Distribution \\ $p_h\uparrow$}   & 
\makecell[c]{Two-tone \\Distribution; \\ Dark Channel Prior \\ $p_h\uparrow$} &
\textcolor{blue}{\textit{Future Direction $\star$}}&
\makecell[c]{Extreme Channel Prior \\ $p_h\uparrow$}
\\
\hline
\makecell[c]{Transformation \\ as Prior} &
\makecell[c]{Wavelet; \\ Edge; \\ $p_l\uparrow;p_h\uparrow$} &
\makecell[c]{Wavelet; \\ $p_l\uparrow;p_h\uparrow$} &
\makecell[c]{Wavelet; \\ $p_l\uparrow;p_h\uparrow$} &
\makecell[c]{Wavelet; \\ $p_l\uparrow;p_h\uparrow$} &
\textcolor{blue}{\textit{Future Direction}} 
\\
\hline

\makecell[c]{Kernel and Noise \\ Information as Prior} &
\makecell[c]{Kernel Modeling; \\ $p_l\uparrow,f_\Phi\uparrow$} &
\makecell[c]{Kernel Modeling; \\ $p_l\uparrow,f_\Phi\uparrow$} &
\textcolor{blue}{\textit{Future Direction} $\star$}&
\makecell[c]{Noise Modeling; \\ $p_l\uparrow,f_\Phi\uparrow$} &
\textcolor{blue}{\textit{Future Direction}}
\\
\hline
\makecell[c]{Semantic \\Information\\ as Prior} &
\makecell[c]{Semantic \\ Segmentation \\ $f_\Phi\uparrow$}   &
\makecell[c]{Semantic Segmentation; \\ Object Detection \\ $f_\Phi\uparrow$}&
\makecell[c]{Semantic \\ Segmentation \\ $f_\Phi\uparrow$} &
\makecell[c]{Semantic \\ Segmentation \\ $f_\Phi\uparrow$} &
\makecell[c]{Semantic \\ Segmentation \\ $p_l\uparrow,f_\Phi\uparrow $}\\
\hline

\makecell[c]{ Non-local \\ Self-similarity \\ as Prior} &
\makecell[c]{ Non-local \\ Self-similarity \\ $f_\Phi\uparrow$}&
\makecell[c]{ Non-local \\ Self-similarity\\ $f_\Phi\uparrow$}&
\textcolor{blue}{\textit{Future Direction} $\star$}&
\makecell[c]{ Non-local \\ Self-similarity\\ $f_\Phi\uparrow$}&
\textcolor{blue}{\textit{Future Direction}}
\\
\hline
Facial Prior &
\makecell[c]{Facial Component;\\ Geometry\\ $f_\Phi\uparrow$} &
\makecell[c]{Facial\\ Component\\ $f_\Phi\uparrow$} &
\textcolor{blue}{\textit{Future Direction}}&
\textcolor{blue}{\textit{Future Direction}$\star$}&
\textcolor{blue}{\textit{Future Direction}$\star$}
\\
\hline
Deep Image Prior 
& \makecell[c]{Deep Image Prior \\ $p_h\uparrow$; $f_\Phi\uparrow$}
& \makecell[c]{Deep Image Prior \\ $p_h\uparrow$; $f_\Phi\uparrow$}
& \makecell[c]{Deep Image Prior \\ $p_h\uparrow$; $f_\Phi\uparrow$}
& \makecell[c]{Deep Image Prior \\ $p_h\uparrow$; $f_\Phi\uparrow$}
& \makecell[c]{Deep Image Prior \\ $p_h\uparrow$; $f_\Phi\uparrow$}
\\
\hline
\makecell[c]{Pre-trained Model\\ as Prior} & \makecell[c]{GAN Inversion; \\ Generative Prior; \\ Deep Denoiser Prior;\\ \makecell[c]{Large-scale \\Foundation Model}\\$p_l\uparrow$;$p_h\uparrow$}& 
\makecell[c]{Generative Prior\\ $p_l\uparrow$;$p_h\uparrow$} &
\makecell[c]{Large-scale \\Foundation Model \\$p_l\uparrow$;$p_h\uparrow$}
&
\makecell[c]{GAN Inversion \\ $p_l\uparrow$;$p_h\uparrow$}&
\makecell[c]{Large-scale \\Foundation Model \\$p_l\uparrow$;$p_h\uparrow$}
\\
\hline
\end{tabular}
}
\centering
\vspace{-7pt}
\end{table*}

\vspace{-10pt}
\section{Open Problems and New Perspectives}
\label{sec:discussion}
\vspace{-5pt}
\subsection{Open Problems}

By far, we have reviewed the technical advancements of prior in various deep image restoration and enhancement tasks.
These priors are closely connected with each other and play pivotal roles in network structures, data generation, and loss function design, \etal.
Therefore, it is essential to understand the connections among priors and their scopes of applications, which are crucial for sparking new research areas in the community. Here, we mainly discuss some critical aspects of prior, including their relationships, generalization capability, potential values, and applications. 

\noindent \textbf{The Relationships Between Priors and DL-based Methods.}
Based on the theoretical analysis in Sec.~\ref{sec:definition}, priors influence deep image restoration and enhancement methods in two ways, data ($p_l$ and $p_h$) and networks ($f_\phi$), as shown in Fig. \ref{tab:summarize_priors}.
From the perspective of data, some priors,~\eg, the prior based on the kernel information (Sec. \ref{subsubsec:kernel_prior_2}), the prior based on the generative models (Sec. \ref{subsubsec:pretrain2}) can increase $p_l$ by utilizing the degradation information in the low-quality (synthesis) images. By contrast, prior related to $p_h$,~\eg, the prior based on the statistical intensity features (Sec. \ref{sec:statistical}), are constructed by learning the intrinsic patterns of high-quality images.
From the perspective of the model, prior can help the convergence of the DNN model and guide its structure design to achieve better performance through increasing $f_\phi$,~\eg, temporal prior (Sec. \ref{subsec:temporal_prior}), semantic prior (Sec. \ref{sec:high_level_information_prior}).
Consequently, it is promising to explore the application of priors in DL-based methods for deep image restoration and enhancement.


\noindent \textbf{The Relationships among Different Priors.}
There are broad connections among priors as many of them reflect the similar properties of images, and some priors can be more detailed representations of others, as summarized in Table~\ref{tab:summarize_priors}. 
For example, dark channel prior~\cite{pan2016blind} and bright channel prior~\cite{yan2017image} are subsets of the prior for channel statistics of images~\cite{yan2017image}.
While extreme channels prior~\cite{yan2017image} and gradient channel prior~\cite{ma2020structure} are concrete representations of image sparsity in different scenes, as discussed in Sec.~\ref{subsubsec:intensity_statistical_feature}.
Temporal sharpness prior~\cite{li2021arvo} is a special form of the temporal prior, as mentioned in Sec.~\ref{subsec:temporal_prior}.
Therefore, it is promising to explore novel priors based on connections among priors. 
For example, the dark channel prior could be extended to adaptive regularization terms to constrain the DNN models in HDR and low-light imaging.

\vspace{-2pt}
\noindent \textbf{Impact of Priors on the Generalization Capacity.}
DNN model training for image restoration and enhancement is purely data-driven; therefore, it tends to find a shortcut of the solution space from the input to output, causing the limited generalization capacity when applied to another data domain~\cite{geirhos2020shortcut}.
A typical example is that many DL-based methods trained on the simulated datasets fail to be generalized well to the more complex real-world scenes~\cite{chen2021progressive}.
Prior is shown to constrain the DNN models better, making them have stronger generalization capacity as prior can facilitate finding a more effective solution space, leading to better generalization performance.
For example, prior can guide the DNNs models of unsupervised domain adaptation to transfer the knowledge from the simulated images to real-world images~\cite{shao2020domain,lengyel2021zero}.
Moreover, it can be a potential direction to transfer knowledge from DNNs trained on simulated datasets generated by different priors to model learning in real-world images.

\noindent \textbf{The Relationship Between Priors and DNN Structures.}
The prior affects the network structure design in deep image restoration and enhancement tasks in three ways.
The first is strongly related to the network itself. For example, non-local attention~\cite{wang2018non} can be the better design to exploit the non-local self-similarity prior~\cite{zhang2019residual} because its structure can fit the property of this prior well.
The second is related to the output. 
Some priors constraint the output of the network, \eg, the dark channel prior~\cite{dharejo2021remote}.
The third is related to the input data. For instance, the optical flow guides the network to use sliding windows or sequences when processing input~\cite{wang2019edvr,chan2021basicvsr++}.
By contrast, the statistical prior (Sec.~\ref{sec:statistical}), \eg, dark channel, and gradient guidance, often act more as the regularization term during training, which is decoupled from the network.
We believe it is imperative and potential to design novel network structures utilizing the property of specific prior.

\noindent \textbf{Beyond Image Restoration and Enhancement.}
Image restoration and enhancement tasks are the basis of almost all high-level vision tasks. Therefore, priors used in deep image restoration and enhancement can potentially be explored to promote the performance of high-level tasks. 
For example, Liu~\etal~\cite{liu2020connecting} used the prior for denoising to improve the performance of deep semantic segmentation models. 
Moreover, facial prior (Sec. \ref{sec:facial_prior}) not only can be used in image enhancement tasks but also can be used in human face detection, segmentation, and recognition~\cite{chen2021progressive}.
Last but not least, the semantic prior, which is commonly used in the high-level and low-level vision tasks (Sec. \ref{sec:high_level_information_prior}), can be connected with other priors to reinforce learning both tasks.
Future research could consider designing more efficient network structures for the high-level and image restoration tasks together because previously both tasks were learned independently.

\noindent \textbf{Why Some Priors are Not Widely Applied to DNNs?}
The first is that DNN itself can automatically learn some features of the images, \eg, blur kernel in deblurring.
For example, Su \etal~\cite{nah2017deep} proposed an end-to-end deblurring network, which outperforms the previous framework\cite{sun2015learning} integrating prior of the explicit blur kernel. 
The second is that some priors, \eg, non-local self-similarity prior (Sec. \ref{sec:non}), have not been well studied so far, thus showing great potential for more active research. 
For example, the recent transformer frameworks can get better results than CNN frameworks in many tasks, such as classification~\cite{khan2021transformers}, and detection~\cite{liu2021swin}, because the transformer framework can get non-local attention.
The last reason is that these priors are used more as the regularization~\cite{chen2019blind} or loss function~\cite{lee2020unsupervised}.
For example, the dark channel prior is designed to a loss function in a deep dehazing network~\cite{golts2019unsupervised}.
In a nutshell, we believe that there exists a large room for research in this direction as the prior could be used in various deep image restoration and enhancement tasks.

\vspace{-18pt}
\subsection{New Perspectives}
\vspace{-5pt}
\label{flag:future_direction}
\label{sec:future_direction}

As most priors indicate the intrinsic property of images, they can be potentially extended to other tasks. 
However, our review discovers some gaps in the applications of priors for specific deep restoration and enhancement tasks, as shown in Tab. \ref{tab:summarize_priors}.


\noindent \textbf{Potential of Priors Based on Kernel Modeling.}
In Sec. \ref{subsec:kernel_prior}, we reviewed and analyzed the priors based on the kernel and noise information for deep image SR and deblurring. Some DL-based image restoration tasks, including dehazing, deraining, and desnowing, rarely use these priors.
However, kernel estimation is an essential step in dehazing, deraining, and desnowing~\cite{wang2020rain}. Consequently, a potential direction is to convert the estimated kernels to a novel type of priors for other deep image restoration and enhancement tasks.

\noindent \textbf{Potential of Non-local Self-similarity Priors.}
Non-local similarity prior reflects the explicit patterns for natural images, as discussed in Sec.~\ref{sec:non}. 
However, Tab.~\ref{tab:summarize_priors} shows that it is employed only in typical image restoration and enhancement tasks, such as image deblurring and SR. This prior exploits the reappearance of some small patches in the image to supply the extra detail for image restoration~\cite{lefkimmiatis2017non}. 
Such detailed information is also helpful for other restoration and enhancement tasks, including HDR imaging, image deraining, image dehazing, etc. Therefore, it is possible to extend this prior to a wider application.

\noindent \textbf{Potential of Facial Priors.}
Research on facial priors mainly focuses on facial hallucination and deblurring. However, the universal features in facial priors should be used for more tasks. For instance, Shen~\etal~\cite{shen2018deep} exploited the geometric and spatial information of the facial components in facial priors for face image hallucination. This geometric and spatial information is common in human faces, which indicates that it might be used to support HDR for face images and low-light image enhancement. Therefore, it is a potential direction to utilize facial priors for HDR imaging, low-light enhancement, deraining, etc, as shown in Tab. \ref{tab:summarize_priors}.

\noindent \textbf{Potential of Statistical Image Features as Priors.}
Some statistical feature priors of the image, \eg, $L_0$ regularization~\cite{pan2016l_0}, gradient profile prior~\cite{sun2008image} and histogram equalization~\cite{dhal2021histogram}, are widely used in traditional image restoration and enhancement tasks.
These priors describe some specific properties of high-quality images or low-quality images, which can be useful in terms of loss function and regularization terms.
For example, the gradient profile prior could be input into the deep SR networks as the edge information, sharpening the edges of the HR images.
Moreover, the histogram equalization prior can also supervise deep HDR networks to predict images with a higher dynamic range.

\noindent \textbf{Priors for Multi-sensor Fusion.}
Recently, event and infrared cameras have brought new inspirations to deep image restoration and enhancement.
Event cameras show many advantages over frame-based cameras, such as HDR and no motion blur \cite{zheng2023deep,lu2023learning}. 
Infrared cameras have also attracted much attention because they can reflect a broader range of wavelengths of light \cite{chudasama2020therisurnet}.
These sensors provide crucial visual information that conventional cameras can not capture in extreme conditions, \eg, high-speed motion and night scenes. 
Endeavors have been made to utilize event data as guidance to boost the performance for deep image restoration and enhancement \cite{wang2021joint}. 
However, research is still urgently needed regarding utilizing these novel camera data as priors to better guide the DNN models for image restoration and enhancement. Moreover, it could be interesting to consider multi-sensor fusion for deep image restoration and enhancement.
We hope that the priors analyzed in this paper can inspire more research to discover new priors for these novel sensor data.

\noindent \textbf{Potential of Pre-trained Models.}
In Sec.~\ref{subsubsec:pretrain1}, some GAN inversion methods~\cite{mGAN_2020_CVPR,deep_generative_prior_dgp} based on the pre-trained generators can generate high-quality images by searching the latent codes in the latent space.
This indicates that the GAN models pre-trained on large-scale datasets can learn some prior knowledge from the high-quality images.
Therefore, developing a task-specific searching strategy for other image restoration tasks is possible.
For instance, for the image dehazing task, the searching targets can be the latent codes of clean scenes while keeping the same semantics with degraded images.
In addition to Sec.~\ref{subsubsec:pretrain2}, some works~\cite{shao2020domain,ye2021closing} propose to use GAN pipelines by jointly embedding the degraded image synthesis and image restoration tasks in an end-to-end manner.
However, these image synthesis methods are limited by the scales of the task-specific training dataset. 
Applying GAN inversion techniques for image synthesis could provide a novel perspective of leveraging the image distribution learned from the pre-trained GAN.
Consequently, it is a promising direction that can further increase the data diversity by generating more unseen images.
More recently, the rapid development of the large-scale pre-trained models shows great potential in various downstream tasks, \eg, zero-shot semantic understanding~\cite{alayrac2022flamingo} and image generation~\cite{saharia2022photorealistic}.
These pre-trained models can serve as strong prior to providing semantic information for various image restoration tasks due to their large-scale training data.

\noindent \textbf{Potential of the Large-scale Foundation Models.}
\label{subsubsec:potential-foundation}
In Sec.~\ref{subsubsec:large-model}, we have summarized the large-scale foundation models as priors, like CLIP~\cite{radford2021learning}, Stable Diffusion~\cite{rombach2022high}, and SAM~\cite{kirillov2023segment}.
Diverse downstream applications~\cite{nichol2021glide,zhou2022learning} have emphasized the versatility of these large-scale foundation models in providing solid semantic representation.
Although some large-scale foundation models have been used in image enhancement and restoration tasks, the substantial emergent capabilities of these models are still far from being fully utilized, such as the instruction-following ability.
For instance, recent large-scale language models, such as GPT-4~\cite{openai2023gpt4}, have shown better zero-shot capabilities than previous models when following the user's instructions.
Looking to the future, the potential of large-scale foundation models in image restoration and enhancement is immense. As demonstrated in Visual ChatGPT~\cite{wu2023visual}, the quality of an image can be iteratively improved by integrating multi-modality input \ie, text, through a user-friendly, interactive interface.
Moreover, the interactive approach of systems like Visual ChatGPT allows for continuous refinements, yielding more precise results based on the text instructions. 
With the development of large-scale foundation models, we anticipate the advanced capacities can be applied to boosting image enhancement and restoration task performance. For instance, harnessing the power of instruction-following for a deeper semantic understanding or task personalizing of images could notably enhance the accuracy and fidelity of image restoration based on user intents.

\vspace{-10pt}
\section{Conclusion}
\vspace{-5pt}
\label{sec:Conclusion}

In this paper, we offered an insightful and methodical
analysis of the recent advances of priors in deep image restoration and
enhancement. 
We conducted a hierarchical and structural taxonomy of prior commonly used DL-based methods. Meanwhile, we provided an insightful discussion on each prior regarding its principle, potential, and applications. Moreover, we summarized the crucial problems by highlighting the potential future directions to spark more research in the community.

\bibliography{references}

\end{document}